\RequirePackage{fix-cm}
\documentclass[twocolumn]{svjour3}          % twocolumn
\smartqed  % flush right qed marks, e.g. at end of proof
\usepackage{graphicx}
\usepackage{times}
\usepackage{amsmath,stackrel}
\usepackage{extarrows}
\usepackage{amsfonts}
\usepackage{booktabs}
\usepackage{algorithm}
\usepackage{algpseudocode}      % layout for algorithmicx
\usepackage{multirow}
\usepackage{subcaption}
\usepackage{float}
\usepackage{xcolor}
\usepackage{hyperref}
\spnewtheorem{assumption}{Assumption}{\bfseries}{\itshape}
 \newcount\DraftStatus  % 0 suppresses notes to selves in text
 \ifnum\DraftStatus=1
 	\usepackage[draft,colorinlistoftodos,color=orange!30]{todonotes}
 \else
 	\usepackage[disable,colorinlistoftodos,color=blue!30]{todonotes}
 \fi 
 \makeatletter
  \providecommand\@dotsep{5}
  \def\listtodoname{List of Todos}
  \def\listoftodos{\@starttoc{tdo}\listtodoname}
  \makeatother
 \usepackage{color}
 
\begin{document}

\title{Deep Treatment-Adaptive Network for Causal Inference%\thanks{Grants or other notes
}


\author{Qian Li        \and
        Zhichao Wang \and
        Shaowu Liu \and 
        Gang Li \and
        Guandong Xu*
        %etc.
}


\institute{Qian Li \at
              School of Electrical Engineering, Computing and Mathematical Sciences, Curtin University, Perth, Australia 
              %\email{qian.li@uts.edu.au}           %  \\
            \and 
            Zhichao Wang \at
               School of Electrical Engineering and Telecommunications, University of New South Wales, Sydney, Australia
             \and
             Shaowu Liu, Guandong Xu \at
               Data Science and Machine Intelligence Lab,  School of Computer Science, University of Technology Sydney, Sydney, Australia \\
              \email{guandong.xu@uts.edu.au}           %  \\
           \and 
           Gang Li \at
           Centre for Cyber Security Research and Innovation, Deakin University,  Geelong, VIC 3216, Australia
}

\date{Received: date / Accepted: date}

\maketitle

\begin{abstract}
 Causal inference is capable of estimating the treatment effect (i.e., the causal effect of \emph{treatment} on the \emph{outcome}) to benefit the decision making in various domains. 
 One  fundamental  challenge  in  this  research  is that the  treatment  assignment bias in observational data. 
 To increase the validity of observational studies on causal inference, representation based methods as the state-of-the-art have demonstrated the superior performance of treatment effect estimation.
 Most representation based methods assume all observed covariates are pre-treatment (i.e., not affected by the treatment), and learn a balanced representation from these observed covariates for estimating treatment effect.
%  Unfortunately, as estimating treatment effect requires to do intervention on treatment, which thus changes the post-treatment covariates as well. 
 Unfortunately, this assumption is often too strict a requirement in practice, as some covariates are changed by doing an intervention on treatment (i.e., post-treatment).
 By contrast, the balanced representation learned from unchanged covariates thus biases the treatment effect estimation.
 %This assumption thus biases the treatment effect estimation, conditioning on different interventions on treatment for estimating treatment effect.
 %place affects on the outcome as well.  
 %This change will further affect the outcome, which results in the bias between the interventional distribution and observed distribution as
 %In other words, representation methods can not distinguish the direct effect from the treatment on the outcome and the mediate effect
 
 %, the covariates changed by the treatment place additional effect on outcome 
%  In addition, 
%  most representation methods follow the traditional machine learning principle
%  by extracting a confounding representation with treatment-shared information, 
%  but neglect the existence of mediate variables 
%  that may bias the treatment effect estimation.
 %Unfortunately,as we have discussed, the current approach of manysocial scientists is to simply condition on posttreatmentvariables—an approach that has the potential to intro-duce serious bias 
 In light of this, we propose a deep treatment-adaptive architecture (DTANet) that can address the post-treatment covariates and provide a unbiased treatment effect estimation. 
 Generally speaking, the contributions of this work are threefold. 
 First, 
 our theoretical results guarantee DTANet can identify treatment effect from observations.
 Second, 
 we introduce a novel regularization of orthogonality projection to ensure that
 the learned confounding representation is invariant and not being contaminated by the treatment, 
 meanwhile mediate variable representation is informative and discriminative for predicting the outcome.  
 Finally, 
 we build on the optimal transport and learn a treatment-invariant representation 
 for the unobserved confounders to alleviate the confounding bias. 
\keywords{Causal inference \and Treatment effect estimation   \and Deep neural networks}
\end{abstract}

\section{Introduction}

%\qian{treatment-invariant->confounding representation; mediator representation}
%1. \qian{That's why we use optimal transport for exploiting the different distribution}
%In the past decades,estimating causal effects from observational study has attracted increasing attention.

Causal inference aims at estimating 
how a treatment affects the outcome~\cite{pearl2009causal,pearl2009causality,li2021causall}, 
which is a common problem in many research fields, 
including medical science~\cite{wager2018estimation}, 
economics~\cite{alaa2017bayesian},
education~\cite{hill2011bayesian}, recommendation~\cite{li2021causal2,schnabel2016recommendations} and statistics ~\cite{bottou2013counterfactual,li2021causal,sun2015causal}.
Taking medical science as an example, 
pharmaceuticals companies have developed many medicines for a certain illness. 
They want to know \emph{which medicine is more effective for a specific patient}.
The treatment effect is defined as the change of the outcome of individuals~\footnote{An ``individual'' can be a physical object, a firm, an individual person, or a collection of objects or persons} 
if an intervention is done on the treatment.
In the above example of medicines,
the individuals could be patients, 
and an intervention would be taking different medicines.
Treatment effect estimation aims to exploit the outcomes 
under different interventions done on the treatment, 
which are necessary to answer the above question and thus it leads to better decision making.

Two types of studies are usually conducted for estimating the treatment effect, 
including the \emph{randomized controlled trials} (RCTs)~\cite{colnet2020causal,concato2000randomized} and observational study~\cite{nichols2007causal,rosenbaum2005observational}. 
RCTs randomly assigns individuals into a treatment group or a control group, 
which is the most effective way of estimating treatment effect.
However, 
randomized controlled trial is often cost prohibitive and time consuming in practice. 
In addition, 
ethical issues largely limits the applications of the randomized controlled trials.
Unlike RCTs, 
observational study becomes a feasible method, 
as it can estimate treatment effect from observational data without controls on the treatment assignment.

\begin{figure*}[]
	\begin{center}
		\includegraphics[width=0.65\textwidth]{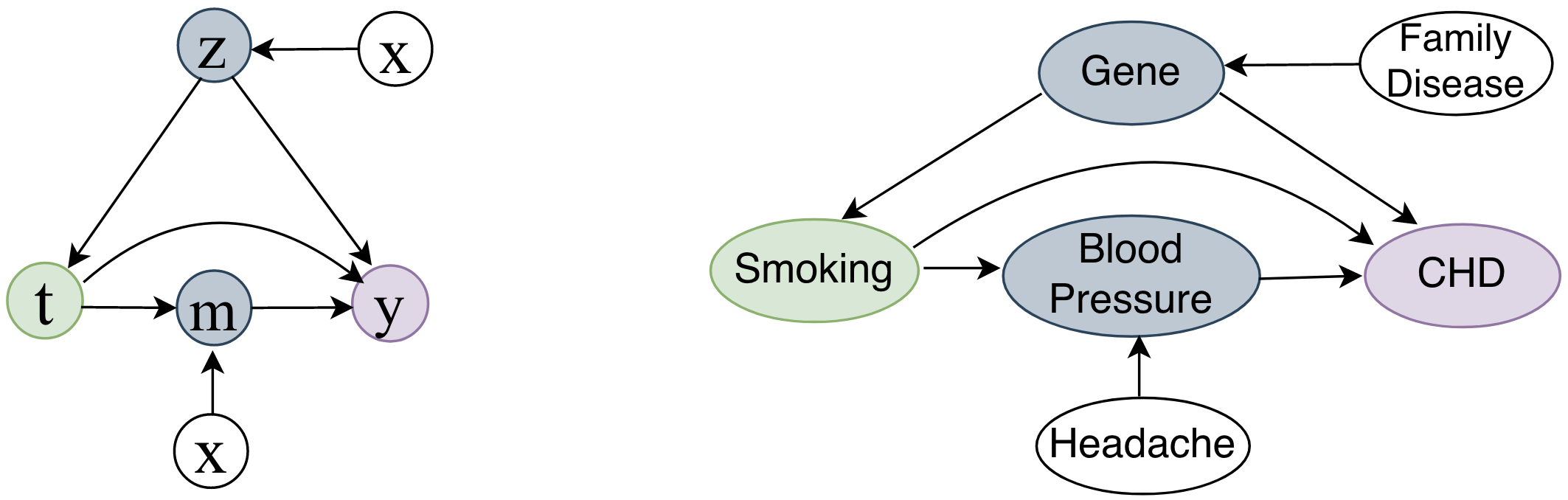}
	\end{center}
	\caption{The causal graph with the mediate variable and its example. 
		The confounder $z$ and the mediate variable $m$ in grey are unmeasured in observational study. 
		We can observe some covariates $x$ that are in fact noisy views of $z$ and $m$, 
		such as the headache and family heart disease.
		}
	\label{fig:cg}
\end{figure*}

Observational studies have attracted increasing attention in the past decades, 
where the hallmark is that the treatment observed in the data 
depend on variables which might also affect the outcome, resulting in confounding bias.
For example in Figure~\ref{fig:cg}, 
we are interested in the effect of treatment \texttt{smoking} on the outcome \texttt{CHD}. 
We have \texttt{gene} causes an individual become more susceptible to \texttt{smoking} 
according to recent studies on genetics of smoking~\cite{davies2009genetics}, 
and specific \texttt{gene} also increases the risk of developing \texttt{coronary} heart disease (CHD).
Moreover, 
the variable \texttt{gene} affects both the treatment \texttt{smoking} and the outcome \texttt{CHD}.
In other words, statistically, we find strong positive association between \emph{Smoking} and \emph{CHD}, which however can be attributed to a causal relationship or/and a spurious correlation resulted from the change in \emph{gene}. 
Consequently, the confounder factors should be untangled, 
otherwise the treatment effect of \texttt{smoking} on \texttt{CHD} is overestimated by the spurious correlation.
The challenge is how to untangle these confounding factors and 
make valid treatment effect estimation~\cite{pearl2009causality,pearl2016causal}.
Causal inference works under the common simplifying assumption of ``no-hidden confounding'', i.e., all counfouders can be observed and measured from observed covariates.
The standard way to account for treatment effect is 
by ``controlling'' the confounders from the observed covariates~\cite{pearl2009causal,pearl2009causality}.  
Particularly, 
confounders lead to the distribution shift that exists 
between groups of individuals receiving different treatments. 
The challenge is how to untangle confounding bias and 
make valid counterfactual predictions 
what if a different treatment had been applied. 
Existing methods for untangling confounders (``controlling'' confounders) 
generally fall into three categories, 
namely propensity-based, 
proxy variable-based, 
and representation based methods. 
Among them,
propensity based methods ``control'' the confounders 
by adjusting representative covariates (e.g., \texttt{age}) that may contain confounding information.
Through this, 
treatment effects can be estimated by direct comparison 
between the treated and the controlled individuals~\cite{rosenbaum1983central,diamond2013genetic}.
These methods are gaining grounds in various applications, 
but a significant challenge is that 
confounders are usually latent in the observational data. 
However, 
such methods require the confounders to 
be measured from observed covariates~\cite{pearl2009causal,pearl2009causality}, 
whereas, in practice, confounders are usually latent in the observational data. 

An alternative to classic method leverages 
the observed ``proxy variables'' in place of unmeasured confounders 
to estimate the treatment effect~\cite{kuroki2014measurement,kallus2018causal}. 
%The proxy is then used in a second stage to estimate the causal effect. 
%The causal effect can be identified if the covariates contain all the confounding factors. 
%For example, we cannot measure the socio-economic status of patients directly, but we might be able to get a proxy by knowing their job type. 
%The proxy is then used in a second stage to estimate the causal effect. 
%The causal effect can be identified if the covariates contain all the confounding factors. 
%For example, we cannot measure the socio-economic status of patients directly, but we might be able to get a proxy by knowing their job type. 
However, 
even with the availability of proxy variables,
the uncertainty of confounder type still makes causal inference a challenge~\cite{louizos2017causal}, 
and thus blocks the accuracy of treatment effect estimation from being improved.  
% To relax the prior assumptions on the confounder, 
% representation based methods learn the features of confounders 
% to estimate the potential outcome more accurately.
The third category has predominantly focused on
learning representations regularized to balance these confounding factors 
by enforcing domain invariance with distributional distances.
%learns the balanced representation for all covarites to adjust the confounder variable. 
Conditioning on the balanced representation, 
the treatment assignment is independent of  confounders,
and thus it alleviates the confounding bias.
The learned feature is balanced across the treated and the controlled individuals 
to alleviate the confounding bias, 
which is guaranteed to be invariant for the different treatment assignments.

%the performance of classical propensity score methods can be easily degraded as the decrease of the overlap between the treated and controlled individuals~\cite{king2019propensity}.
%Ignorance of the confounder will overestimate or underestimate the effect of the treatment according to the positive or negative confounder.

Although deep representation based methods have  shown superior performance for causal inference, 
they still suffer from two significant drawbacks. 
First,
the learned representation ignores treatment-specific variations affected by different treatments, 
which results in biased treatment effect estimation.
This assumption is too strong and invalid in practice, 
as some covariates are usually changed after doing intervention on the treatment. 
This leads to the bias to treatment effect estimation, 
as it requires to compute between the interventional distribution and observed distribution.
These post-treatment covariates are frequently observed in practice.
By acting as mediate variables, 
post-treatment covariates can place effects on outcomes and treatment effect estimation.
\iffalse
Most representation based methods can only address the simple case with no unobserved confounders, 
which however is in practical for causal inference. 
Other causal inference methods resort to estimate the unobserved confounders 
by defining the special causal variables, 
which however requires typical prior causal knowledge.
\fi
%Previous representation methods focus only on learning a confounding representation, ignoring the unique characteristics of different treatment
%On the one hand, 

A typical example is that smoking can cause \texttt{coronary heart disease (CHD)} through increasing the \texttt{blood pressure (BP)}, 
as indicated in Figure~\ref{fig:cg}. 
The \texttt{blood pressure} involving treatment-specific variations 
is called a mediate variable, 
that may vary under the different treatments. 
Thus, 
simply using treatment indicator will lose significant information for the outcome prediction 
and thus lead to biased treatment estimation.
The causal relationships among treatment, 
 mediate feature and  outcome 
are largely unexploited in previous representation based methods. 
%two individuals with the same treatment may have different outcomes, i.e., one has CHD and one does not, due to the different blood pressures response to the  treatment. 
%On the other hand, treatment indicator might get lost if the learned representation is high-dimensional~\cite{shalit2017estimating}. 
%Second, previous representation methods assume that all covariates are pre-treatment variables and the learned representation is invariant to different treatments, which violates the fact that some covariates can be changed by intervention on the treatment. 
%because the dimension of treatment (e.g., one-dimension for the binary settings) is far less than the dimension of confounding representation, as the dimension increases~\cite{shalit2017estimating}.  
In addition, 
some covariates (\texttt{blood pressure}) may be changed 
by doing an intervention on treatment (\texttt{smoke behaviour}) 
and are usually neglected by previous representation methods.
Previous representation methods fail to learn the individual characteristics of each group. We argue that explicitly modeling what is unique to each group can improve a model’s ability to extract treatment–invariant features and thus benefit for estimating unbiased treatment effect.

In this work, 
we propose an end-to-end \emph{deep treatment-adaptive network} (DTANet) 
to estimate the treatment effect as shown in Figure~\ref{fig:fc}.
To the best of our knowledge, 
the proposed DTANet is the first representation based method 
that can quantify the mediate effect transmitted by the change of treatment.
%to capitalize on the dependency structure among the confounder, treatment, outcome and mediation .we propose a deep model for causal inference based on the treatment adaption,in the presence of unknown confounder.Different from existing representation based methods,our method explicitly models the mediating factor as the treatment-specific features to improve the causal inference ability.  
%Rather than simply deriving the balanced representation of confounding, our model separately models the confounder bias and treatment-adaptive features

%that are adaptive to different treatments in the observational data, 

\begin{itemize}
	\item  
	By a novel orthogonality projection, 
	a mediate feature representation can be learnt to 
	capture the informative treatment-specific variations underlying the unobserved mediate variables.
	%allows to capture the unique  within a specific treatment
	%As the intervention on the treatment assignment may break the treatment-invariant principle of confounders, our model proposes 
	The mediate feature representation independent of 
	unobserved confounders can generate an unbiased estimation of the mediate treatment effect.
	
	\item Our DTANet leverages the optimal transport theory to 
	learn a treatment-invariant representation that can alleviate the confounders bias.
	%We consider the fact that some covariates may be post-treatment variables
	%The learned representation exploits the shared confounding information by reducing Wasserstein distance between the distribution of treated and controlled individuals. 
	Moreover, 
	the learned treatment-invariant features can be employed 
	as the off-the-shelf knowledge in estimating  causal effect on out-of-samples.
	%to guarantee the generalization for treatment effect estimation.
	%Applying intervention on the treatment may result in different covariates distributions. The learned confounding representation can reduce the discrepancy between different treatments .

	%reduce the noise  influence of treatment on the confounding separate the treatment-specific representation from the confounding representation to guarantee the generalization for treatment effect estimation.
	%is not affected by the confounding representation is invariant to the change of the intervention on the treatment.
	
	%By partitioning the space in such a manner, the classifier trained on the shared representation is better able to generalize across domains as its inputs are uncontaminated with aspects of the representation that are unique to each domain.
	
	\item Finally, 
	DTANet is an end-to-end deep joint network with two separate  ``heads'' for two potential outcomes, 
	by using both the confounding representation 
	and the mediate feature representation.
	We also prove that the causal effect can be identified from the observational data by DTANet. 
	%Under the mild assumption，we can prove that the causal effect maybe computed from observational data.
\end{itemize}

\section{Background}
	
	This section introduces the preliminary knowledge and related work in the field of observational studies.
	\subsection{The Rationality of Causal Inference}
	The goal of causal inference is to estimate the causal effect of an intervention/treatment.
    Randomized controlled trials (RCTs) are now the gold standard for causal inference in medicine and social science. In RCTs, individuals are receiving treatment or controlled treatment by randomization.
    RCTs allow to estimate the treatment effect by directly comparing the results from assigning the intervention of interest to the results from a ``control'' intervention. 
    For example, researcher in medicine are interested in assessing the effect of smoking on the health outcome.
    % The main benefit of the RCT is the use of a randomisation procedure that, 
    RCTs assign individuals randomly with smoking and non-smoking.
    Due to randomization and given a large enough study enrollment, the two study groups (smoking and non-smoking) are fully comparable.  
    That means they will have roughly the same number of individuals at baseline and the same number of individuals in each age (or gender/occupation/etc.) group. The only differences between the two groups should be due to the assignment, all other things (e.g., gender, age, occupation, etc.) having been made equal. Therefore, a direct comparison between two groups' average health outcome is thus a valid effect estimation of the smoking vs. non-smoking.
    % We might compare the health outcomes for people taking smoking with the health outcomes for nonsmokers.
    However, performing RCTs would be neither feasible in behavioral and social science research due to practical or ethical barriers. 
    Because it is impossible to assign people chosen at random to smoke for decades.

    Observational studies (or non RCTs) that do not impose any intervention of the individuals' treatment resort to purely observational data. Unlike the randomized control trials, the mechanism of treatment assignment in observational studies is not explicit.
    For example, instead of randomized experiments, individuals take smoke based on several factors rather than being assigned randomly. 
    As a result, the distribution of smoking group will generally be different from the non-smoking group.
    A direct comparison between the health outcomes for smokers and the health outcomes for nonsmokers is no longer valid for estimating the effect of smoking on health outcomes.
    In this situation, causal inference that is capable of estimating causal effects from observational study are of paramount importance.
% 	Estimating individual treatment effect from observational data faces two major challenges, missing counterfactuals and treatment selection bias.
    
	\subsection{Potential Outcome Framework}
\label{sec:pof}
    Two well-known fundamental causal paradigms, including the potential outcome framework~\cite{rosenbaum1983central} and structural causal models~\cite{pearl2009causal,pearl2016causal}, are adopted in causal inference from observational studies. In this paper, we focus on the potential outcome framework.
	%\qliMarker
	The potential outcome framework~\cite{rosenbaum1983central} proposed by Neyman and Rubin has developed into a well-known causal paradigm for treatment effect estimation in observational studies.
	Considering binary treatments for a set of individuals, there are two possible outcomes for each individual.
	In general, the potential outcome framework predicts counterfactual (i.e., outcome under an alternative treatment) for each treated individual, and computes the difference between the counterfactual and the factual (observed outcome). 
	
	Formally, for an observational dataset $\{\boldsymbol{x}_i,t_i,y_i\}_{1\leq i\leq n}$ of $n$ individuals,  variable $\boldsymbol{x}_i\in\mathbb{R}^{n\times d}$ 
	is the $d$-dimensional covariate of individual $i$, and
	treatment $t_i$ affects the outcome $y_i$.
	Considering the binary treatment case, individual $i$ will be assigned to the control group if $t_i=0$, or to the treated group if $t_i=1$. 
	The \emph{individual treatment effect} (ITE) is defined as the difference between potential outcomes of an individual under two different treatments:
	\begin{equation}
	\centering
	\operatorname{ITE}_i=\mathbb{E}(\mathbf{y}_i(1))-\mathbb{E}(\mathbf{y}_i(0))
	\label{eq:ite_o}
	\end{equation}
	Clearly, each individual only belongs to one of these two groups, and therefore, we can only observe one of two possible outcomes. In particular, if individual $i$ is in treated group, $y_i(1)$ is the observed/factual outcome, and $y_i(0)$ is missing data, i.e., counterfactual.
	The challenge to estimate ITE lies on how to estimate the missing counterfactual outcome $y_i(0)$ by intervening $t=0$.
	%$t\in {0,1}$ indicating that the individual receives treated treatment or controlled treatment. 
	%According to the Rubin-Neyman causal model~\cite{rubin2005causal},we assume two potential outcomes $y_0$ and $y_1$.

% 	Under these assumptions, 
% 	the intervention distribution is non parametrically identified as
% 	\begin{equation}
% 	\begin{split}
% 	p(y\mid x,do(t=1))&=\mathbb{E}[p(y\mid x,do(t=1),z)]\\
% 	&=\int_z p(y\mid x,t=1,z)p(z) dz
% 	\end{split}
% 	\label{eq:intv}
% 	\end{equation}
% 	Based on Eq.~\eqref{eq:intv}, 
% 	we can compute ITE by estimating $p(y\mid x,t=1,z)$ from observational studies.

	The potential outcome framework usually makes the following assumptions~\cite{imbens2000role,lechner2001identification} to estimate the missing counterfactual outcome.
	\begin{assumption}
		[\textbf{Ignorability}]
		Conditional on the covariates $\boldsymbol{x}$, 
		two potential outcomes are independent of the treatment, i.e. $y_i(1), y_i(0) \perp t \mid  \boldsymbol{x}$.
	\end{assumption}
	%Given the observed pretreatment confounders, the treatment assignment is assumed to be ignorable, that is, statistically independent of potential outcomes and potential mediators. In such situations, a common strategy of empirical researchers is to collect as many pretreatment confounders as possible so that the ignorability of treatment assignment is more
	\begin{assumption}
		[\textbf{Positivity}]
		For any set of covariates $\boldsymbol{x}$, 
		the probability of receiving each treatment $a$ is positive, 
		i.e., $0<p(t=a\mid x)<1$.
	\end{assumption}
	Estimating causal effects from observational data is different from classic learning 
	because we never see the ground-truth individual-level effect in practice. 
	For each individual, we only see their response to one of the possible actions - the one they had actually received.

% 	\begin{equation}
% 	\operatorname{ITE}(x)=\mathbb{E}(\mathbf{y}( \boldsymbol{x}, d o(t=1)))-\mathbb{E}(\mathbf{y}(\boldsymbol{x}, d o(t=0)))
% 	\label{eq:ite_o}
% 	\end{equation}

	%Applying the do-operator allows it to make specific interventional predictions about events within the causal model. 
	%\iffalse
	%According to the Rubin-Neyman causal model~\cite{rubin2005causal}, two potential outcomes exist for $\boldsymbol{x}$ with treatments $\{0,1\}$, respectively.  
	%\begin{equation}
	%\begin{split}
	%y_0(\boldsymbol{x})=\mathbb{E}(\mathbf{y}( \boldsymbol{x}, d o(t=1))),\quad
	%y_1(\boldsymbol{x})=\mathbb{E}(\mathbf{y}(\boldsymbol{x}, d o(t=0)))
	%\end{split}
	%\end{equation}
	%\begin{equation}
	%\operatorname{ITE}(x)=\mathbb{E}[\mathbf{y} | \boldsymbol{x}, d o(t=1)]-\mathbb{E}[\mathbf{y} | \boldsymbol{x}, d o(t=0)]
	%\label{eq:}
	%\end{equation}
	%\fi
	%Importantly, we only observe the component of the potential outcome vector that corresponds to the assigned treatment, we call this the factualoutcome, and refer to unobserved potential outcomes as counterfactual outcomes or justcounterfactuals. 

% 	\begin{figure}[!htb]
% 		\begin{center}
% 			\includegraphics[width=0.25\textwidth]{fig/do.png}
% 		\end{center}
% 		\caption{An illustration of $do$-operator: $do$ operator on $t$ will remove influence from other variables to $t$, i.e., the dashed line can be deleted.}
% 		%$t$ is the treatment, $Y$ is the outcome and $z$ is the common cause of the treatment and outcome, i.e., confounder.
% 		\label{fig:do}
% 	\end{figure}

	\subsection{Confounders and Bias}
	\label{subsec:cb}

	The problem of calculating ITE is translated into 
	the task of estimating the counterfactual outcome under an intervention on treatment.
	Hence, the potential outcome framework introduces a mathematical operator called $do$-calculus $do(t)$ 
	to define hypothetical intervention on the treatment $t$~\cite{pearl2009causality}.  
	Specifically, 
	$do(t)=1$ simulates an intervention by setting $t=1$, 
	which indicates that 
	$t$ is only determined by $do$ thus renders $t$ independent of the other variables.
	\begin{definition}[Interventional Distribution]
	    The interventional distribution $p(y\mid do(t'))$ denotes the distribution of the variable $y$ when we rerun the modified data-generation process where the value of variable $t$ is set to $t'$.
	\end{definition}
	For example, for the causal graph in Figure~\ref{fig:cg}, the post-intervention distribution $p(y\mid do(0))$ refers to the distribution of \texttt{CHD} outcome $y$ as if the smoking treatment $t$ is set to $0$ (e.g. non-smoking) by intervention, where all the arrows into $t$ are removed.
	However, 
	the interventional distribution $p(y\mid do(t'))$ is different from observational distribution $p(y\mid t')$ due to the existence of confounders. 
	\begin{definition}[Confounders]
	    Given a pair of treatment and outcome $(t,y)$, 
	    we say a variable $z$ is a confounder iff $z$ affects both $t$ and $y$.
	\end{definition}
	Confounder is a common causes of the treatment and outcome.
	The confounder variable affects the assignment of individuals' treatment and thus leads to the confounding bias. In the medicine example, \texttt{gene} is a confounder variable, so that people with different
   \texttt{gene} have different preferences on smoking or not.
	The probability distribution $p(y|t)$ not only includes the effect of treatment on the outcome (i.e., $p(y\mid do(t))$), but also includes the statistical associations produced by confounders on the outcome, 
	which leads to the spurious effect. 
	Consequently, 
	confounders render the probability distribution $p(y\mid t)$ and intervention distribution $p(y\mid do(t))$ distinct, 
	which makes calculating ITE more difficult.
% 	Confounders lead to confounding bias, which makes the counterfactual outcome estimation more difficult.
% 	This gap can give rise to confounding bias, which results if one estimates treatment effects using $p(y\mid t)$ where $p(y\mid do(t))$ is in fact required.
	\begin{definition}[Confounding Bias]
	    Given variables $x$, $y$, confounding bias exists for causal effect $t\rightarrow y$ iff the observational probabilistic distribution is not always equivalent to the interventional distribution, i.e., $p(y\mid t)\neq p(y\mid do(t))$.
	\end{definition}
	Confounding bias in observational study is equivalent to a domain adaptation scenario where a model is trained on a ``source'' (observed) data distribution, but should perform well on a ``target'' (counterfactual) one.
	 Handing confounding bias is the essential part of causal inference, and the procedure of handing confounder variables is called \emph{adjust confounders}.

	\section{Related work}

	Estimation of individual treatment effect in observational data is a complicated task due to the challenges of confounding bias~\cite{xu2020causality,pearl2009causality,dung2021stochastic}. 
	Unlike the randomized control trials, the mechanism of treatment assignment is not explicit in observational data due to the confounding bias. Therefore, interventions of treatment are not independent of the property of the subjects, which results in the difference between the intervention (i.e., counterfactual) distribution and the observed distribution. To predict counterfactual outcomes from the factual data, many practical solutions are proposed to adjust confounders, which can be classified into four categories.

	% Since the separate model does not account for confounding bias.
% 	Due to the existence of confounders, the covariate distributions of the treated group and control group are different, which leads to the confounding bias problem .
	A common statistical solution is re-weighting certain data instances to balance the observed distribution and intervention distributions caused by confounding bias problem (as described in Section~\ref{subsec:cb}). 
	Apparently, confounding bias leads to the fact that treatment assignment is not random but is correlated with covariates.
	By defining an appropriate weight as the function of covariates to each individual in the observational data, a pseudo-population can be created on which the distributions of the treated group and control group are similar.
	In other words, the treatment assignment is synthesized to be random after weighting individuals.
	The majority of re-weighting approaches belong to the \emph{Inverse Propensity Weighting} (IPS) family of methods~\cite{austin2011introduction}.
	Here,
	the propensity denotes the estimated probability of receiving a treatment~\cite{rosenbaum1983central}, which is often modelled by a logistic regression of treatment on the covariates. 
	IPS weights the individuals with inverse propensity to make a synthetic random treatment assignment and further create unbiased estimators of treatment effect.
Methods in the second category is matching, 
	which provides a way to estimate the counterfactual while reducing the confounding bias brought by the confounders.
% 	estimates the counterfactual outcome of an individual with respect to treatment using the factual outcomes of its nearest neighbours that received an alternative treatment
	According to the (binary) treatment assignments, a set of individuals can be divided into a treatment group and a control group. For each treated individual, matching methods select its counterpart in the control group based on certain criteria, and treat the selected individual as a counterfactual. 
	Then the treatment effect can be estimated by comparing the outcomes of treated individuals and the corresponding selected counterfactuals. Various distance metrics have been adopted to compare the closeness between individuals and select counterparts.
	Some popular matching estimators include \emph{nearest neighbor matching} (NNM)~\cite{rubin1973matching}, propensity score matching~\cite{rosenbaum1983central}, and genetic matching~\cite{diamond2013genetic}, etc.
	In detail, a propensity score measures the propensity of individuals to receive treatment given the information available in the covariates. 
	In Figure \ref{fig:cg}, we can estimate the propensity score by fitting a logistic model for the probability of quitting smoking conditional on the covariates.
	Propensity score methods match each treated individual to 
	the controlled individual(s) with the similar propensity score 
	(e.g., one-to-one or one-to-many), 
	and then treat the matched individual(s) as the controlled outcome~\cite{diamond2013genetic,bang2005doubly}.
	The individual treatment effect equals to 
	the difference between the matched pair of the treated individual and the controlled individual.

	Methods in the third category learn individualized treatment effects (ITE) via parametric regression models to 
	exploit the correlations among the covariates, treatment  and outcome. 
	\emph{Bayesian Additive Regression Trees} (BART)~\cite{hill2011bayesian}, 
	\emph{Causal Random Forest} (CF)~\cite{wager2018estimation} 
	and \emph{Treatment-Agnostic Representation Network} (TARNet)~\cite{shalit2017estimating} are typical methods of this category.
	In particular, 
	BART in~\cite{hill2011bayesian} applies a Bayesian form of boosted regression trees 
	on covariates and treatment for estimating ITE, and it is capable of addressing non-linear settings and obtain more accurate ITE 
	than the propensity score matching and 
	inverse probability of weighting estimators~\cite{hill2011bayesian}.
	\emph{Causal random forest} (CF) views forests as a adaptive neighbourhood metric and 
	estimates the treatment effects at the leaf node~\cite{wager2018estimation}. 
	TARNet~\cite{shalit2017estimating} is a complex deep model that 
	builds on learning non-linear representations 
	between the covariates and potential outcomes.
	\emph{Doubly Robust Linear Regression} (DR)~\cite{dudik2011doubly} combines the propensity score weighting with the outcome regression, so that the estimator is robust even when one of the propensity scores or outcome regression is incorrect (but not both).

% 	In the second, the treatment is considered a feature, with one model learned for everything, and the mismatch between the entire sample distribution and treated and control distributions is adjusted in order to account for confounding bias. 
%     Representation based methods of the third category 
% 	considers a feature, with one model learned for everything, 

    The fourth category has predominantly focused on learning representations regularized to balance these confounding factors by enforcing domain invariance with distributional distances~\cite{johansson2016learning,schwab2018perfect}. The big challenge in treatment effect estimation is that the intervention distribution is not identical to the observed distribution, which converts the causal inference problem to a domain adaptation problem~\cite{li2021hilbert,li2015lingo}. Building on this work~\cite{johansson2016learning}, the discrepancy distance between distributions is tailored to adaptation problems. An intuitive idea is to enforce the similarity between the distributions of different treatment groups in the representation space.
	%On top of the balanced representation, the regression model (e.g., ridge-regression ) is fit for predicting the outcomes.
	Two common discrepancy metrics in this area are used: 
	empirical discrepancy by \emph{Balancing Neural Network} (BNN)~\cite{johansson2016learning} 
	and maximum mean discrepancy by \emph{Counterfactual Factual Regression Network} (CFRNet)~\cite{shalit2017estimating}.
	Particularly, BNN learns a balanced representation that adjusts the mismatch between the entire sample distribution and treated and control distributions in order to account for confounding bias.
	CFRNet provides an intuitive generalization-error bound.
	The expected ITE representation error is bounded by the generalization-error and the distribution distance.
	The drawback of methods in this category is that 
	they overlooks the important information that can be estimated from data: 
	the treatment/domain assignment probabilities~\cite{johansson2018learning}. 
	\iffalse
	Following this idea, a few improved models have been proposed and discussed. For example, ~\cite{johansson2018learning} brings together shift-invariant representation learning and re-weighting methods. ~\cite{hassanpour2019counterfactual} presents a new context-aware weighting scheme based on the importance sampling technique, on top of representation learning, to alleviate the confounding bias problem in ITE estimation.
	\fi

% 	Perfect Match (PM) develops the performance metrics and a deep model architecture for estimating individual treatment effects 
% 	in the setting with multiple treatments~\cite{schwab2018perfect}.

%\gliMarker %TODO: GLi here

		\section{Problem Formulation}
	
	%the mediate variable (mediating) and confounder are denoted as $M$ and $z$. 
	%We use $\mathcal{M}$, $\mathcal{z}$, $\mathcal{x}$ and $\mathcal{Y}$, respectively. 
	%Consider a simple random sample of size $n$ from non-experimental data, 
	%we observe $(t_i, \boldsymbol{x}_i,y_i)$ for each individual $i$ in~\eqref{fig:cg}, 
	%where $\boldsymbol{x}_i$ denotes the vector of the observed covariates, 
	
	%Similarly, we use $y_i(t_i, m)$ to represent the potential outcome for individual $i$ when $t_i=t$ and $m_i=m$. 
	
	\subsection{Motivation} 
	Treatment can cause the outcome directly or indirectly through mediation (e.g., blood pressure). 
	The indirect cause is largely unexploited by most of the previous representation methods, 
	which leads to the biased estimation of treatment effect.
	In this paper, 
	we consider the causal graph in Figure~\ref{fig:cg} with confounder and mediate variable. 
	Both the confounder and the mediate variable may not be amenable to direct measurements. 
	It is reasonable to assume that both the confounder and the mediate variable can be reliably represented by a set of covariates for each individual.
	For example,
	even if the family gene and blood pressure can not be measured directly, 
	they can also be reflected by the family disease and the headache as shown in Figure~\ref{fig:cg}. 
	We will prove that true treatment effect in Figure~\ref{fig:cg} 
	can be identified from observations by our DTANet.
	%The identification of such proxy variables relies on the strong prior causal knowledge, which however is domain-specific and not easy to obtain. To alleviate this issue, our method can learn treatment-adaptation features
	%without prohibitive assumptions on the unobserved confounders or the mediate variables.

	%some confounder measurements may be contaminated with noise (e.g., data recording error), while other confounders may  and instead admit only proxy measurements.
	%Perhaps more practically, 
	
	%The treatment $T$ and outcome $Y$ are commonly caused by confounder $z$ (e.g., by family gen),which is unobserved and can be approximated by the proxy variables underlying the covariates. 

	%The identification of such intermediate mechanism is essential for generating explanations and counterfactual analysis must be invoked to facilitate this identification.
	
	\subsection{Theoretical Results}
	\label{sec:tre}
	
	We admit the existence of mediate variable and consider the causal graph in Figure~\ref{fig:cg}. 
	Next, we define the potential outcomes. Previously, the potential outcomes were only a function of the treatment, but in our scenario the potential outcomes depend on the mediate variable as well as the treatment variable.
	Assume $m(t_i)$ is the mediate variable under the treatment status $t_i$, and $z$ is the unobserved confounder. 
	The mediate variable is a post-treatment variable and can be changed by the intervention on treatment. 
	This change will further affect the outcome, 
	which results in the bias between the interventional distribution and observed distribution as
	\begin{equation}
	p(y_i\mid do(t=1),m_i(t),x_i)\neq p(y_i\mid t=1,m_i,x_i)
	\end{equation} 
	In this case, 
	the bias will lead to invalid ITE in Eq.~\eqref{eq:ite_o}. 
	Consequently, 
	extracting the mediate variable from the covariates is vital for the unbiased the treatment effect estimation.

	%Traditional ITE~\eqref{eq:ite_o} merely exploits the changes of outcomes under the intervention of treatment and keeps the remaining covariates invariant.
	
	Our goal is to estimate ITE under the existence of mediate variable. 
	We reformulated ITE defined in Eq.~\eqref{eq:ite_o} as Eq.~\eqref{eq:ite} and prove that it is be identified from observations.
	%$p(y_i\mid do(t=1),m_i(1),x_i)$ is 
	%identify the causal effect 
	
	\begin{equation}
	\begin{split}
	\tau_{ITE}(\boldsymbol{x})=\mathbb{E}[ y(t, m(t))\mid \boldsymbol{x},do(t=1)]\\
	-\mathbb{E}[ y(t, m(t))\mid \boldsymbol{x},do(t=0)]
	%\\&=\underbrace{y_{i}\left(1, M(0)\right)-y_{i}(0, M(0))}_{\text{nature direct effect}}+\underbrace{y_{i}\left(1, M(1)\right)-y_{i}\left(1, M(0)\right)}_{\text{indirect effect bias  }\Delta_i}
	%\\\stackrel[]{(a)}{=}& \frac{1}{2}\sum_{t=0}^1\left(\delta(t_i)+\zeta(1-t)\right)\\
	%\delta(t_i)= & \mathbb{E}(y_{i}\left(t, M(1)\right))-\mathbb{E}(y_{i}\left(t, M(0)\right))\\
	%\zeta(t_i)= &\mathbb{E}(y_{i}\left(1, M(t)\right))-\mathbb{E}(y_{i}\left(0, M(t)\right))
	%\delta(t_i)&\equiv y_{i}\left(t, M(1)\right)-y_{i}\left(t, M(0)\right)\\
	%\zeta(1-t)&\equiv y_{i}\left(1, M(1-t)\right)-y_{i}\left(0, M(1-t)\right)
	\label{eq:ite}
	\end{split}
	\end{equation}

	\begin{theorem}
		The causal effect defined by \emph{ITE} in Eq.~\eqref{eq:ite} can be identified from the distribution $p(\boldsymbol{x},t,y)$.
        \label{thm:ice}
	\end{theorem}
	\begin{proof}
		%Apparently, 
		\emph{ITE} can be non-parametrically identified by  
		\begin{equation}
		\begin{split}
		&p(y(t,m(t)) \mid  \boldsymbol{x}, d o(t=1))\\&=
		\int_{m} p(\mathbf{y} \mid  \boldsymbol{x}, m) p(m \mid  \boldsymbol{x}, d o(t=1)) d m\\& \stackrel[]{(i)}{=}\int_{m} p(\mathbf{y} \mid  \boldsymbol{x}, m) p(m \mid  \boldsymbol{x}, t=1) d m\\&=\int_{m}\int_{z}p(\mathbf{y} \mid  \boldsymbol{x}, z, m) p(z\mid \boldsymbol{x},m)p(m \mid  \boldsymbol{x}, t=1) d md z\\& \stackrel[]{(ii)}{=}\int_{m}\int_{z}p(\mathbf{y} \mid  z, m) p(z\mid \boldsymbol{x})p(m \mid  \boldsymbol{x}, t=1) d md z
		%\\&=\int_{\mathbf{U}}\int_{z} \int_{t'}p(\mathbf{y}\mid t',\mathbf{U},z)p(t') p(z\mid \mathbf{X})p(\mathbf{U} \mid  \mathbf{X}, t) dt'dzd \mathbf{U}
		%\\& \stackrel[]{(ii)}{=}\sum_{t}\int_{\mathbf{U}} p(\mathbf{y} \mid  \mathbf{X}, d o(\mathbf{U})) p(\mathbf{U} \mid  \mathbf{X}, d o(t)) d \mathbf{U} 
		\label{eq:do}
		\end{split}
		\end{equation}
		%From the derivations above, it is apparent that the key to employing natural direct and indirect effects is to identify and estimate the distribution, or aspects of the distribution, of the nested counterfactuals Y (a, M (a*)) for potentially different a and a*. 
		%If one is only interested in a given function g and contrasts such as E[g (Y (1, M (1)))] − E[g (Y (0, M (1)))]
		%If the observed covariates $X$ include all common causes of the treatment and outcome—i.e., block all backdoor paths— then the causal effect is equal to a parameter of the observational distribution $p(X,t,y)$.
		%we need to block the back-door paths to alleviate the confounder bias or irrelevant causal effects that are not from treatment. 
		
		%The case for $t=1$ is identical, and 
		%which can be readily recovered from the probability function $p(\mathbf{y} \mid  \mathbf{X}=x, d o(t=0))$ and $p(\mathbf{y} \mid  \mathbf{X}=x, d o(t=1))$. 
		%ATE is known if \emph{ITE} is computed. 
		%We have that:
		%Since for all $r \in \mathcal{R}$, $t \in\{0,1\}$, we have 
		%\begin{equation}
		%\begin{split}
		%    p_{\Phi}(t \mid  x)&=p(t \mid  \Psi(x)) \\ 
		%    p\left(Y_{t} \mid  x\right)&=p\left(Y_{t} \mid  \Psi(x)\right)
		%\end{split}
		%\end{equation}Fig.~\ref{fig:fc}
		According to Figure~\ref{fig:cg}, 
		there is no common cause between the treatment and the mediate variable. 
		Therefore, 
		the interventional distribution $p(m\mid \boldsymbol{x},do(t=1))$ 
		equals the observed distribution $p(m\mid \boldsymbol{x},t=1)$, 
		which allows equality (i) in Eq.~\eqref{eq:do} to be satisfied. 
		As indicated by Figure~\ref{fig:cg}, 
		when the confounder $z$ is conditioned, 
		$y$ is independent of $x$, i.e., $y\perp \boldsymbol{x}\mid z$. 
		Similarly, 
		$z$ is independent of $m$ when $\boldsymbol{x}$ is conditioned, i.e., $z\perp m\mid \boldsymbol{x}$. 
		The equality (ii) holds because of $y\perp \boldsymbol{x}\mid z$ and  $z\perp m\mid \boldsymbol{x}$.
		The final expression only depends on the distribution $p(\boldsymbol{x},z,m,t,y)$. 
		%Then, substituting~~\ref{eq:exp} into the definition of $\delta(t_i)$ and $\zeta(1-t)$ in~~\ref{eq:\emph{ITE}} yields the desired results for \emph{ITE}.
		%The result for the average \emph{ITE} can be also computed. 
		Similarly, 
		we can also prove that 
		$p(\mathbf{y}(t,m(t)) \mid  \boldsymbol{x}, d o(t=1))$ can be expressed 
		by observations $p(\boldsymbol{x},z,m,t,y)$. 
		Based on \emph{ITE} in Eq.~\eqref{eq:ite},
		we can conclude that \emph{ITE} can be computed 
		by recovering the distribution $p(\boldsymbol{x},z,m,t,y)$ from the observational dataset $(\boldsymbol{x},t,y)$. 
		
		%In practice, this requires estimating the unknown confounder $z$ and mediate variable $m$.
		%means that there is no other confounding bias for the causal effect of $t$ on $U$, i.e., $\mathbf{X}$ blocks the back-door paths from $U$ to $Y$ according to back-door criterion~\cite{}. 
	\end{proof}
	
	\subsection{Representation Learning for $z$ and $m$}
	\label{s:r}
\begin{figure}[!hbt]
		\begin{center}
			\includegraphics[width=0.45\textwidth]{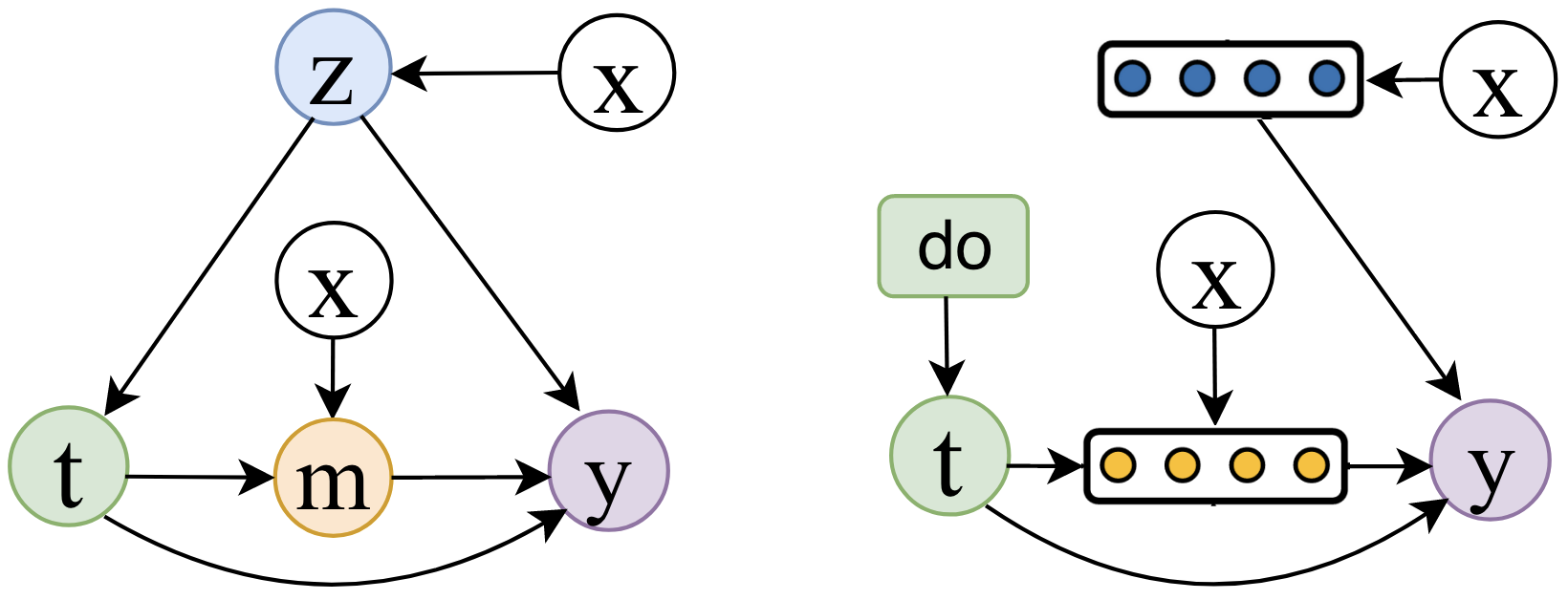}
		\end{center}
		\caption{The representation based causal graph for unobserved confounder $z$ and mediate variable $m$. }
		\label{fig:cm}
	\end{figure}

	Identification of treatment effects relies on causal assumptions, which can be encoded in a causal graph. This is the fundamental assumption for causal inference methods. 
	In this paper, 
	we design a representation based causal graph shown in Figure~\ref{fig:cm}, based on which we propose deep treatment-adaptive network (DTANet) for treatment effect estimation.
	Our method is based on the same causal graph that is widely used by previous causal inference methods, i.e., $(T\leftarrow Z\rightarrow Y, T\rightarrow Y)$. In addition, we extend this causal graph by involving the existence of $m$ between $t$ and $y$.
	DTANet learns the latent confounding representation and the mediate feature representation for the unmeasured confounders $z$ and mediate variables $m$,
	respectively.
	As proved in theorem~\ref{thm:ice}, 
	conditioning on the $z$ and $m$ would amplify the treatment effect estimation bias. 
	Defining proxy variables for unmeasured $z$ and $m$ requires 
	domain-specific prior knowledge that is not easy to obtain.
	Consequently, 
	our task is to learn two latent representations to 
	filter out the information related to $z$ and $m$ from covariates, 
	which requires no prohibitive assumption or knowledge on unobserved $z$ and $m$. 
	
	%we propose a novel model to derive the representations for $z$ and $m$, 
	%In the next section, we will show how to learn the representation so as to estimate treatment effect in Eq~\eqref{eq:ite}.

	\textbf{Debiasing confounder $z$.} 
    The confounding representation is learned from covariates with the aim of alleviating the confounding bias. The treatment assignment is not randomly but typically biased by the confounder. For example, poor patients are more likely to choose the cheap treatment, where the economic status as a confounder determines the choice of treatment. The distribution of individuals may therefore differ significantly between the treated group and the overall population. A supervised model naïvely trained to minimise the factual error would overfit to the properties of the treated group, and thus not generalise well to the entire population.
	
	According to theorem~\ref{thm:ice}, 
	inferring causal effect would be straightforward if the confounder $z$ is available. 
	So, 
	as the substitute for the unknown confounder, 
	we would like to learn a treatment-invariant representation from the observed covariates. 
	We justify the rationality of this strategy based on: 
	1) as the confounder is hidden in the observable covariates, i.e., 
	the family gene is hidden in the family disease, 
	confounder can be learned from covariates; 
	2) as $do$-calculus removes the dependence of treatment on confounder shown in Figure~\ref{fig:cm}, 
	the substitution of the confounder should capture
	the generalized or mutual information of covariates, i.e., treatment-invariant property. 
	The learned representation with treatment-invariant property containing the covariate features such that the induced distributions of individuals under different treatments look similar, which can thus generalize well to the entire population.

	\textbf{Mediate feature learning for $m$.} 
	Previous representation based models neglect the interactions 
	between the treatment and the individuals' covariates, 
	i.e., doing different interventions on the treatment may result in varied mediate treatment effects
	that can further change the observed covariates as well. 
	Neglecting such change in the observed covariates will lead to 
	serious bias for the treatment effect estimation, 
	as the confounding representation is learned from the static covariates. 
	Namely, some covariates are in fact mediate variables that can be changed by a different treatment value.
	To capture the dynamic changes private to different treatments, 
	we learn a mediate feature representation of unobserved mediate variables.

	%It is reasonable because mediator is a post-treatment variable that occurs before  the  outcome  is  realized.

	\subsection{Causal Quantities of Interest}
	The treatment effect can be measured at the individual level and group level.
	\subsubsection{Individual Level}
	The key quantity of interest in causal inference is treatment effect on outcome. Based on \emph{ITE} in Eq.~\eqref{eq:ite} and Theorem~\ref{thm:ice}, we have \emph{ITE} for each individual $i$ as
    \begin{equation}
        \tau_{{ITE}_i}=   y_{i}(1, m_{i}(1),\boldsymbol{x}_i)-y_{i}(0, m_{i}(0),\boldsymbol{x}_i)
        \label{eq:ite2}
    \end{equation}
    where $y_{i}(1, m_{i}(1),\boldsymbol{x}_i)$ is the treated outcome of individual $i$ after applying $do(t_i)=1$, $m_{i}(1)$ is the mediate variable resulting from $do(t_i)=1$ and $\boldsymbol{x}_i$ is the covariate vector. Similar to treated outcome, $y_{i}(0, m_{i}(0),\boldsymbol{x}_i)$ is the controlled outcome after applying $do(t_i)=0$.

	%\textbf{Mediate Treatment Effect.}
% 	The superior of our DTANet lies in the analysis of mediate variable for treatment effect estimation. 
    % \emmph{Mediate Effect Effect} quantifies the effect of treatment on outcome that occurs through a mediate variable. 
	We define the \emph{Mediate Treatment Effect (MTE)} to quantify the effect of treatment on outcome that occurs through a mediate variable. 
	\begin{equation}
	    \tau_{{MTE}_i(t)}= y_{i}(t, m_{i}(1))-y_{i}(t, m_{i}(0))
	    \label{eq:me1}
	\end{equation}
	Note that $\tau_{MTE}$ is computed by applying $do$-calculus on $m$ and keeping $t$ unchanged. 
	The key to understanding Eq.~\eqref{eq:me1} is the following counterfactual question: What change would occur to the outcome if one changes  $m$ from $m(0)$ to $m(1)$, while holding the treatment status at $t$? 
	If the treatment $t$ has no effect on the $m$ , that is, $m(0)\neq m(1)$, then the mediate treatment effect is zero.
	%Although Yi(t, Mi(t)) is observable for units with Ti  t, Yi(t, Mi(1  t)) can never be observed for any unit.
% 	\begin{equation}
% 	    \tau_{MTE}(i)=y_{i}\left(t, m_{i}(1),\boldsymbol{x}_i\right)-y_{i}\left(t, m_{i}(0),\boldsymbol{x}_i\right)
% 	\end{equation}
% 	For $\tau_{ITE}$, we apply $do$-calculus on treatment $t$, resulting in the changes in $m$ as well. 
%     At population level, the \emph{Average Mediate Effect (AME)} for each individual $i$ is defined by
% 	This causal quantity is the change in the outcome corresponding to a change in the mediate variable from the treated value  ($t=1$) to the controlled value ($t=0$).

	%\textbf{Direct Treatment Effect.} 
	We also are interested in \emph{Direct Treatment Effect} that computes how much of the treatment variable $t$ directly affects the outcome $y$.
	Similarly, we can define the individual direct effect of the treatment as follows:
	\begin{equation}
    \tau_{{DTE}_i (t)}=  y_{i}(1, m_{i}(t))-y_{i}(0, m_{i}(t))
    \label{eq:dte}
    \end{equation}
    which denotes the direct causal effect of the treatment on the outcome other than the one represented by the mediate variable. Here, the mediate variable is held constant at $m_{i}(t)$ and the treatment variable is changed from zero to one.

    Finally, the sum of~\eqref{eq:me1} and~\eqref{eq:dte} equals~\eqref{eq:ite2}, which
formally decomposes \emph{ITE} into \emph{Mediate Treatment Effect} and \emph{Direct Treatment Effect} as follows.
    \begin{equation}
        \tau_{{ITE}_i}=\tau_{{MTE}_i(t)}+\tau_{{DTE}_i(1-t)}
    \end{equation}

	\subsubsection{Population Level}
	Given these individual-level causal quantities of interest, we can define the population average effect for each quantity.
	At the population level, the individual treatment effect is named as the \emph{Average Treatment Effect (ATE)}, which is defined as:
    \begin{equation}
    \begin{split}
        \tau_{ATE}&=  \frac{1}{n}\sum_{i}^n \left( y_{i}(1, m_{i}(1))-y_{i}(0, m_{i}(0))\right)\\
        &=\frac{1}{n}\sum_{i}^n\tau_{{ITE}_i}
    \end{split}
    \end{equation}
    Suppose we have $n_t$ treated individuals, \emph{Average Treatment effect on the Treated group (ATT)} is defined as
    \begin{equation}
	\tau_{ATT}=\frac{1}{n_t}\sum_i^{n_t}\tau_{ITE}(i|t=1)
	\end{equation}
	where $n_t$ is the number of individuals having $t=1$, i.e., the treated group size. Here, $\tau_{ITE}(i|t=1)$ is \emph{ITE} for individual $i$ from the treated group. 
	
	Similarly, we define average \emph{Mediate Treatment Effect} and \emph{Direct Treatment Effect} as
	\begin{equation}
	    \tau_{AME}=\frac{1}{n}\sum_i^{n} \tau_{MTE}(i),\quad \tau_{ADE}=\frac{1}{n}\sum_i^{n} \tau_{DTE}(i)
	\end{equation}

% 	We also define the \emph{Average Treatment Effect} as , i.e., the causal effect of the treatment on the outcome through the mediate variable.
% 	\begin{equation}
%     \tau_{DTE} = \frac{1}{2n}\sum_{i}^n \left( y_{i}(1, m_{i}(1))-y_{i}(0, m_{i}(0))\right)
%     \end{equation}

% 	Similarly, at the population level, we define \emph{Average Mediate Effect} as
% 	\begin{equation}
% 	    \tau_{AME}=\frac{1}{n}\sum_i^{n_t} \tau_{IME}(i)
% 	\end{equation}
	
    % \begin{equation}
    %     \tau_{ATE}=  \frac{1}{n}\sum_{i}^n \left( y_{i}(1, m_{i}(1))-y_{i}(0, m_{i}(0))\right)=\frac{1}{n}\sum_{i}^n\tau_{{ITE}_i}
    % \end{equation}

    % \begin{equation}
    %     \zeta_{i}(t) \equiv y_{i}(1, M_{i}(t))-y_{i}(0, M_{i}(t))
    % \end{equation}
	\section{Methodology}
\begin{figure*}[!thb]
		\begin{center}
			\includegraphics[width=0.94\textwidth]{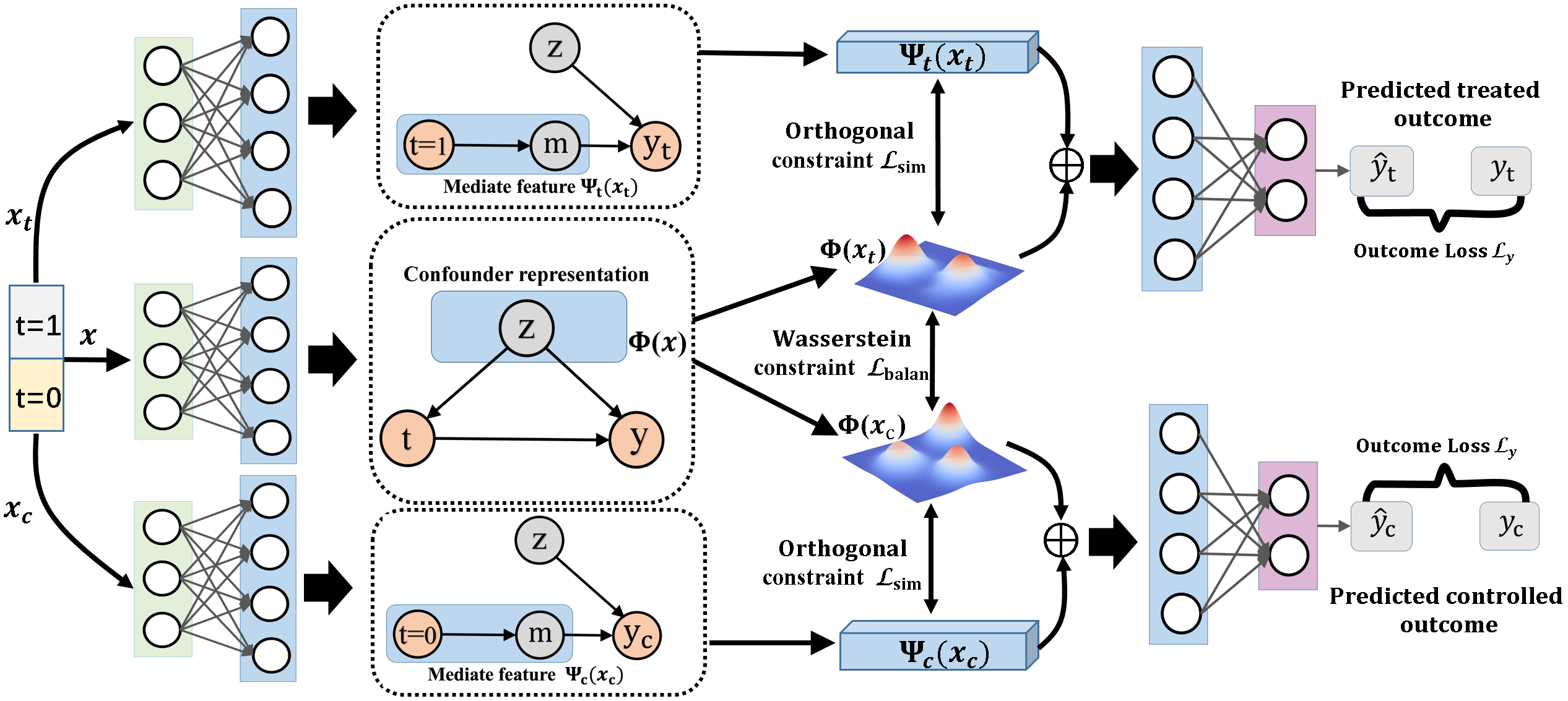}
		\end{center}
		\caption{Our DTANet method provides an end-to-end procedure 
			for predicting potential outcomes from covariates $\boldsymbol{x}$, 
			which can be further used for estimating treatment effect. 
			A confounding representation network $\Phi(\cdot)$, two mediate feature representation networks ($\Psi_t(\cdot)$ and $\Psi_c(\cdot)$) 
			and two predictors of potential outcomes together form DTANet. 
		}
		\label{fig:fc}
	\end{figure*}

	%In this section, we first define some necessary notations that will be used throughout this paper. 
	%The most important assumptions are in the supplement.
	%After that, we will introduce how to estimate causal effects by the proposed architecture in ~\eqref{fig:cg}, and justify the rationality of this strategy.
	%thought this paper and also introduce some necessary ideas about the statistical estimation of causal effects.
	In this section, we learn the representations for unmeasured $z$ and $m$ given in Figure~\ref{fig:cm} in order to compute the \emph{individual treatment effect (ITE)} of Eq.~\eqref{eq:ite}.
	We propose a novel deep treatment-adaptive network (DTANet) as shown in Figure~\ref{fig:fc}. Particularly, DTANet can jointly learn the unbiased confounding representation for $z$ by the optimal transport. Moreover, the mediate features of $m$ viewed as treatment-specific variations can be guaranteed by the proposed orthogonal projection constraint.
	%Note that DTANet is jointly trained for accurately characterizing the causal feature space by confounding representation and treatment-specific representation.
	The confounding representation is concatenated with mediate feature representation 
	for the potential outcome predictor network. 
	With two potential outcomes, 
	the \emph{individual treatment effect (ITE)} can be estimated by Eq.~\eqref{eq:ite}.
	%Note that $\Phi(\boldsymbol{x})$ is parametrized by deep neural networks trained jointly in an end-to-end way, as in~\eqref{fig:fc}.

	%Given the observed covarites of individuals, the goal of causal feature learning is to learn both discriminate and balanced representations w.r.t. treatment-invariant features and treatment-specific features.
	%On the top of the representations of causal features, a non-linear hypothesis model is built for predicting counterfactual outcomes.
	%This allows for learning complex non-linear representations and hypotheses with large flexibility.  

	%The flexible non-linear nature of neural nets will hopefully allow us to approximate well the true interactions between the treatments and its effect.
	
	\subsection{Debiasing Confounder by Optimal Transport}

	%the confounder is the common cause of treatment and outcome. Thus, the substitution for the confounder should contain the most informative features that is generalized and invariant among different treatments for predicting outcome.
	%from the covariates 
	%between $x$ and $(y,t)$ such that $x\perp t,y\mid z$. 
	%Namely, the confounder allows the covariates $\boldsymbol{x}_t$ in the treated group ($t=1$) covariates $\boldsymbol{x}_c$ in the treated group ($t=0$)
	%the confounder can be learned by a low-dimensional representation from covariates, as some covariates can be viewed as the noisy proxies,
	Motivated by the intuition in Section~\ref{s:r}, 
	we define $z=\Phi(;W):\mathcal{X}\rightarrow\mathcal{Z}$ as the representation network 
	for the common confounding information between the treated individuals and the controlled individuals.
	The network $\Phi(;W)$ has $L$ layers with weight parameters $W$ by
	\begin{equation}
	\Phi(\boldsymbol{x};W)=f_L(\ldots f_1(w_{(1)}^{\top} \boldsymbol{x})\ldots) 
	\label{eq:phi}
	\end{equation}
	%\begin{equation}
	%    \Phi(x)=M_{l} \circ \rho_{l-1} \circ M_{l-1} \circ \cdots \circ \rho_{1} \circ M_{1}(\boldsymbol{x})\\
	%    %W_i\Phi(x)_{i-1}
	%    \label{eq:phi}
	%\end{equation}
	where $f_{1}\cdots f_{L}$ are nonlinear activation functions, 
	$w_{(1)}^{\top} x$ is an affine transformation map controlled by weight parameters $w_{1}$ for first layer, 
	and $W=\{w_{(1)},\cdots,w_{(L)}\}$ is the weight matrix for $L$-th layers. 
	%Note that weight parameters also include the bias term for simplicity of notation.

	According to the binary treatment setting, 
	an individual in the observational dataset can be either a treated or controlled individual. 
	To allow $\Phi$ to satisfy the treatment-invariant property, 
	we adopt the optimal transport~\cite{villani2008optimal,peyre2019computational,li2021hilbert,courty2014domain,wang2019polynomial} to minimize the discrepancy introduced by $\Phi$ 
	between the distribution of treated and controlled individuals.
	We use $\boldsymbol{x}_{t}$ for the treated covariates and 
	$\boldsymbol{x}_{c}$ for the controlled covariates. 
	$p(\Phi(\boldsymbol{x}_t))$ and $q(\Phi(\boldsymbol{x}_c))$ are the treated and the controlled distribution induced by $\Phi(\cdot)$. 
	We resort to optimal transport theory  that  allows  to  use Wasserstein distance~\cite{peyre2019computational}
	on the space of probability measures $p(\Phi(\boldsymbol{x}_t))$  and $q(\Phi(\boldsymbol{x}_c))$. Wasserstein metric incorporates the underlying geometry between outcomes, which can be applied to distributions with non-overlapping supports, and has good out-of-sample performance~\cite{esfahani2018data}.
	We apply the Wasserstein distance to reduce the discrepancy even with limited or no overlap 
	between $p(\Phi(\boldsymbol{x}_t))$ and $q(\Phi(\boldsymbol{x}_c))$.

	%Specifically, $\Phi(\cdot)$ is parametrized by a matrix $W$. Apparently,$\Phi(\boldsymbol{x};W)$ maps the covariates $\boldsymbol{x}$ to a treatment-invariant  confounding representation that are common across the treated individuals and controlled individuals.

	%Using this notation, we can write $\Phi(x)$ as a function depending on $W=\{w_{(1)},\cdots,w_{(L)}\}$, i.e., $\Phi(x;W)$. 
	%$\Phi(x)_0 = x$, $\Phi(x)_{i}$ is the state of the $i$-th hidden layer and $\Phi(x)_l$ corresponds to the output of
	%cause effect on treatment both for the treated group ($t=1$) and controlled group ($t=0$). 
	%The causal effect can be identified if the covariates contain all confounding variables. Note that $\Phi(X)$ is parameterized by deep neural networks trained jointly in an end-to-end way, as in~\eqref{fig:fc}. This model allows for 
	%Inspired by, the covariate difference between treated  
	%We motivated $\Phi$ with the intuition that we should use only the information in the confounders that is relevant to both the treatment assignment and outcome. 
	%where $i$-th hidden layer is defined by the weight matrix $W_i$  
	
	%\qian{why balance}

	%As the confounder are the common causes for the treatment assignment no matter $t=1$ or $t=0$
	%We say that a representation $Z_{\Phi}= \Phi(X)$ of $X$ is treatment-invariant if , where 

	\begin{definition}
		Given a hypothesis set $\mathcal{H}$, 
		the Wasserstein distance between $p_{\Phi}$ and $q_{\Phi}$ is
		\begin{equation}
		\mathcal{W}_{2}(p_{\Phi}, q_{\Phi})=\left(\inf _{\pi \in \Pi } \int_{\Omega} d\left(\Phi(\boldsymbol{x}_{t}), \Phi(\boldsymbol{x}_{c})\right)  d \pi\right)^{\frac{1}{2}}
		\label{eq:wp}
		\end{equation}
		where set $\Pi$ is the  joint probability measures on 
		$\Omega=\Phi(\boldsymbol{x}_{t})\times \Phi(\boldsymbol{x}_{c})$ with marginal probabilities $p_{\Phi}$ and $q_{\Phi}$.
	\end{definition}
	
	%We denote $\mathcal{P}_t$ as the set of all probability measures $\mu$ on representation space $\Phi(\cdot)$ of treatment group $\boldsymbol{x}_t$. Similarly, we have the probability measure $\nu\in \mathcal{P}_c$ on the control group $\boldsymbol{x}_c$. 
	As both $p_{\Phi}$ and $q_{\Phi}$  have finite supports, 
	we will only consider Wasserstein distance for discrete distributions. 
	%empirical measures throughout this paper. 
	Given realizations $\{\boldsymbol{x}_{t_i}\}_{i=1}^{n_t}$ and $\{\boldsymbol{x}_{c_j}\}_{j=1}^{n_c}$,
	we reformulate Eq.~\eqref{eq:wp} on two discrete empirical distributions $p_{\Phi}$ and $q_{\Phi}$ 
	w.r.t. treatment individuals and control individuals, i.e., 
	\begin{equation}
	p_{\Phi}=\frac{1}{n_c}\sum_{i=1}^{n_c} \delta_i, \quad q_{\Phi}=\frac{1}{n_t}\sum_{j=1}^{n_t} \delta_j
	\label{eq:ps}
	\end{equation}
	Minimizing the discrepancy between $p_{\Phi}$ and $q_{\Phi}$ 
	with Wasserstein distance is equivalent to solving the optimization
	%reformulate Wasserstein distance $W_p(\mu_0,\mu_1)$ between $p_{\Phi}$ and $q_{\Phi}$ as
	\begin{equation}
	\mathcal{W}_{2}(p_{\Phi}, q_{\Phi}):\stackrel{\text { def }}{=} \min _{\boldsymbol{\gamma}\in \mathbf{U}}\left\langle \mathbf{C}_{\Phi},\boldsymbol{\gamma}\right\rangle_F
	\label{eq:wtc}
	\end{equation}
	where $\langle\cdot,\cdot\rangle_F$ is the Frobenius dot-product of matrices. 
	The optimal $\boldsymbol{\gamma}$ belongs to \begin{equation}
	\mathbf{U} =\left\{\boldsymbol{\gamma}\in \mathbb{R}_{+}^{n_c \times n_t} \mid  \boldsymbol{\gamma}\boldsymbol{1}_{n_t}=p_{\Phi}, \boldsymbol{\gamma}^{\top} \boldsymbol{1}_{n_c}=q_{\Phi}\right\}
	\label{eq:pt}
	\end{equation}
	that refers to non-negative matrices such that 
	their row and column marginals are equal to $p_{\Phi}$ and $q_{\Phi}$ respectively. 
	The distance matrix between $\boldsymbol{x}_t$ and $\boldsymbol{x}_c$ is $\mathbf{C}_{\Phi}\in\mathbb{R}^{n_c\times n_t}$ with  element
	\begin{equation}
	\mathbf{C}_{ij}=\| \Phi(\boldsymbol{x}_{c_i};W)-\Phi(\boldsymbol{x}_{t_j};W)\| ^2_2
	\label{eq:cm}
	\end{equation}
	%where $p_{\Phi}(\boldsymbol{x}_{t_i})$ is the probability of the treatment-invariant feature on the $i$-th treated individual, and $p_{\Phi}(\boldsymbol{x}_{c_i}) $ is the probability of the treatment-invariant features on the $j$-th controlled individual. 
	\iffalse
	\begin{equation}
	\begin{split}
	p_{\Phi}(\boldsymbol{x}_{t})&=\frac{p_{\Phi}(\Phi(x),t=1)}{p(t=1)}=\frac{p(\Phi^{-1}\Phi(x),t=1)J_{\Phi^{-1}}(\Phi(x))}{p(t=1)}\\
	&=p(x\mid t=1)J_{\Phi^{-1}}(\Phi(x))
	\end{split}
	\end{equation}
	\fi
	%\begin{lemma} For $\Phi(X)\in\mathcal{R}$, we have
	% \begin{equation}
	%     p_{\Phi}^{t=1}(\Phi(X))=p(X\mid t=1)
	% \end{equation}
	% \label{lm:p}
	%\end{lemma}
	%the Wasserstein distance in empirical distributtion requires solving the linear program~\cite{}.
	%It is reasonable to define $p_i=p(X\mid t=0)=\frac{1}{n_c}$ and $q_i=q(X\mid t=1)=\frac{1}{n_t}$.

	Hence, 
	we propose Eq.~\eqref{eq:wtc} as the loss $\mathcal{L}_{balan}$ 
	that reduces the discrepancy between the treated and control individuals, i.e.,
	\begin{equation}
	\mathcal{L}_{balan}=\min _{\boldsymbol{\gamma}\in \mathbf{U}}\left\langle \mathbf{C}_{\Phi},\boldsymbol{\gamma}\right\rangle_F
	\label{eq:wd}
	\end{equation}
	%. The representation $\Phi$ allow us to compresses the the relevant information of treated and control distributions into the low-dimensional representation space. 
	Solving $\mathcal{L}_{balan}$ ensures
	the treatment-invariant representation $\Phi$ is similar across different treatment values and thus is independent of the treatment assignment.
	The confounding representation provides more stable gradients even if 
	two distributions of treated and controlled individuals are distant, 
	as well as informative for treatment effect estimation.
	Moreover, 
	since treatment-invariant features are independent of the treatment assignment,
	they can be considered as  off-the-shelf knowledge and used to estimate causal effect on out-of-samples.

	%Define $p_{\Phi}$ define a distribution induced by $\Phi$ over $\mathcal{R}$.

	%For this strategy to work, we learned the embedding by minimizing an objective function $\mathcal{L}_{sim}$. Even if we knew the confounders $U$ we would not actually use all the information they contain to infer the causal effect.  Instead, if we use estimator $\hat{\phi}$ to estimate the effect $\phi$, then we only require the part of $U$ that is actually used by the estimator $\hat{\phi}$. What this means is that if we can build a good predictive model for the treatment then we can plug the outputs into a causal effect estimate directly, without any need to learn the true confounder $U$. The same idea applies generally by using a predictive model for both the treatment and outcome. Reducing the causal inference problem to a predictive problem is the crux of this paper.

	\subsection{Orthogonal Projection for Mediate Features Learning}
	
	%\qian{more emphasized here as our contribution, refer to Deep Counterfactual Networks with Propensity-Dropout}
	%Predicting disease, $Y$ , from risk factors like genetic predisposition or smoking, $X_C$ , and symptoms, $X_E$: while we might have (possibly unlabelled) data from multiple geographical regions or demographic individualss leading to different distributions over risk factors $(D\leftarrow X_C )$, we would not necessarily expect this to affect the behaviour of the disease itself $(X_C\leftarrow y\leftarrow X_E)$.
	According to the binary treatment assignments, 
	individuals in the observational dataset can be either divided into the treated individuals or the controlled individuals.
	We design two mediate feature representations encoding 
	different treatment-specific variations private to both populations (i.e., the treated individuals and the controlled individuals). 
	Moreover, 
	the confounder is no long correlated with the treatment after $do$ intervention as shown in causal graph Figure~\ref{fig:fc}.
	Thus, 
	a soft orthogonal projection term is also proposed to 
	separate the mediate features from the confounding representation as much as possible. 
	This guarantees 
	the confounding representation is pure and not contaminated by treatment. 

	Similar to  representation by Eq.~\eqref{eq:phi}, 
	let functions $\Psi(\boldsymbol{x}_t;V_t)$ and $\Psi(\boldsymbol{x}_c;V_c)$ 
	map  treated individuals $\boldsymbol{x}_t$ and controlled individuals $\boldsymbol{x}_t$ 
	to  hidden mediate representations specialised in each domain. 
	\begin{equation}
	\begin{split}
	\Psi_t(\boldsymbol{x}_t;V_t)=f_L(\ldots f_1(v_{t_{(1)}}^{\top} \boldsymbol{x}_t )\ldots), \vspace{1pt} \\
	\Psi_c(\boldsymbol{x}_c;V_c)=f_L(\ldots f_1(v_{c_{(1)}}^{\top} \boldsymbol{x}_c )\ldots), \vspace{1pt}  
	\end{split}
	\label{eq:psi}
	\end{equation}
	where $V_t=[v_{t_{(1)}}\cdots v_{t_{(L)}}]$ and $V_c=[v_{c_{(1)}}\cdots v_{c_{(L)}}]$ 
	are weight matrices for $L$-layers of the treated and controlled representation, respectively.
	%$\Psi(x)$ as a function depending on $V=\{v_1,\cdots,v_L\}$, i.e., $\Psi(x,t;V)$. 
	
	%Inspired by~\cite{bousmalis2016domain,salzmann2010factorized}, 
	We propose an orthogonality constraint for the loss $\mathcal{L}_{sim}$ 
	to separate the confounding representation from mediate representation.
	Let $Z_t$ and $Z_c$ be matrices whose rows are the outputs of confounding representation $\Phi(\cdot)$ 
	from treated $\boldsymbol{x_t}$ and controlled individuals $\boldsymbol{x_c}$, respectively. 
	Similarly, 
	let $M_t$ and $M_c$ be matrices whose rows are the outputs of the mediate feature representation
	$\Psi_t(\boldsymbol{\cdot})$  and $\Psi_c(\boldsymbol{\cdot})$, respectively. 
	Mathematically, we have 
	\begin{equation}
	\begin{split}
	\mathcal{L}_{sim}
	%&=\mathcal{L}_{t}(\Phi(x_t),\Psi_{1}(x_t))+\mathcal{L}_{c}(\Phi(x_c),\Psi_{0}(x_c))\\
	&=\|M_t^{\top}Z_t\|_F^2+\|M_c^{\top}Z_c\|_F^2
	\end{split}
	\label{eq:orth}
	\end{equation}
	where $\|\cdot\|_F^2$ is the squared Frobenius norm. 
	The loss function $\mathcal{L}_{sim}$ encourages $\Psi_t(\cdot)$ and $\Psi_c(\cdot)$ 
	to encode discriminative features that are specific to their own domain. 
	As $\Psi_t(\cdot)$ and $\Psi_c(\cdot)$ are deduced by the specific treatment, 
	$\Phi(\cdot)$ is constrained to be as general as possible irrespective of the treatment information.

	\subsection{Joint Two-headed Networks for Outcome Prediction}
	
	%\qian{concentration of two vectors, refer to causal information bottleneck}
	Parametrizing two potential outcomes with a single network as in~\cite{johansson2016learning} is not optimal,
	because the influence of $t$ on the potential outcome might be too minor to lost 
	during the training for the high-dimensional case of $\Phi$. 
	%The potential outcome network is trained separately to estimate two potential outcomes $y_i^{t=1}$ and $y_i^{t=0}$.
	We construct two separate  ``heads'' of the deep joint network $\hat{y}_t$ and $\hat{y}_c$
	for the two potential outcomes under treatment and control, 
	as indicated in Figure~\ref{fig:fc}. 
	The concatenation of $[\Phi(\cdot),\Psi_t(\cdot) ]$ or $[\Phi(\cdot),\Psi_c(\cdot)] $ 
	is ultimately fed into the potential outcome network $\hat{y}_t$ or $\hat{y}_c$, 
	respectively. 
	Namely, each sample is used to update only the head corresponding to  observed treatment.
	%The input of $Y^{t}$ is the concatenation of $\Phi(x_t)$ and $\Psi(x_t,1)$.
	%$h_1(\Phi) =h(\Phi,\Psi(X,1))$ and $h_0(\Phi) =h(\Phi,\Psi(X,0))$. 
	%We parameterize causal features representation $\Phi,\Psi$ and outcome prediction function $h(\Phi,\Psi)$ by deep neural networks trained jointly, 
	
	\begin{equation}
	\begin{aligned} 
	%\tilde{s}(x) &=f\left(\ldots f\left(\mathbf{r}_{s}^{(1)} \odot\left(\mathbf{w}_{s}^{(1)}\right)^{\top} x\right) \ldots\right) \\ 
	\hat{y}_{t}(\Phi,\Psi_t;\Theta_t) &=f_L(\ldots f_1( \theta_t^{\top}(\Phi(\boldsymbol{x}_t), \Psi_t(\boldsymbol{x}_t))) \ldots)\\ 
	\hat{y}_{c}(\Phi,\Psi_c;\Theta_c) &=f_L(\ldots f_1(\theta_c^{\top}(\Phi(\boldsymbol{x}_c), \Psi_c(\boldsymbol{x}_{c}))) \ldots)\\ 
	%\mathcal{L}_{y}&=\lambda_1\sum_{i=1}^{n_t}L(\hat{y}_{t}-\hat{y}_t)+\lambda_0\sum_{i=1}^{n_c}L(\hat{y}_{c}-y_c)\\
	\end{aligned}
	\label{eq:y}
	\end{equation}
	where $\Theta_t=[\theta_{t_{(1)}}\cdots \theta_{t_{(L)}}]$ and $\Theta_c=[\theta_{c_{(1)}}\cdots \theta_{c_{(L)}}]$ are weight matrices for $L$ layers of the treated and the controlled, $f_1(\cdot)$ is the first layer
	with the linear transformation weight $\theta_t$ or $\theta_c$ for the treated group or the controlled group, 
	respectively.
	Minimizing the loss function $\mathcal{L}_{y}$ 
	to approximate two predicted potential outcomes to the ground-truths. 
	\begin{equation}
	\begin{aligned} 
	%\tilde{s}(x) &=f\left(\ldots f\left(\mathbf{r}_{s}^{(1)} \odot\left(\mathbf{w}_{s}^{(1)}\right)^{\top} x\right) \ldots\right) \\ 
	%\hat{y}_{t} &=f\left( \theta_t^{\top}(\Phi(\boldsymbol{x}_t), \Psi(\boldsymbol{x}_t,1))\right) \\ 
	%\hat{y}_{c} &=f\left(\theta_c^{\top}(\Phi(\boldsymbol{x}_c), \Psi(\boldsymbol{x}_c,0))\right)  \\ 
	\mathcal{L}_{y}&=\frac{\lambda_0}{n_t}\sum_{i=1}^{n_t}\|\hat{y}_{t_i}-y_{t_i}\|_2^2+\frac{1-\lambda_0}{n_c}\sum_{j=1}^{n_c}\|\hat{y}_{c_j}-y_{c_j}\|_2^2\\
	\end{aligned}
	\end{equation}
	where $\lambda_0$ is a hyper parameter compensating for the difference 
	between the sizes of treated samples and controlled samples.
	%weighting the contribution of each sample to the loss function.
	With the fitted models $\hat{y}_{t}$ and $\hat{y}_{c}$ parametrized by $\Phi,\Psi_t$ and $\Psi_c$ in hand, 
	we can estimate the \emph{individual treatment effect (ITE)} as 
	\begin{equation}
	    \tau_{ITE}(i) =\hat{y}_{t_i}-\hat{y}_{c_i}
	    \label{eq:y}
	\end{equation}
	%Then, the training objective corresponds to adapting the network parameter $(W,V_0,V_1,\Theta)$ so as to minimizing the expected loss $\mathcal{L}_{y}(h(x,t),y)$ as
	\textbf{Remark.} The \emph{mediate feature learning} component enables our approach to estimate the mediate treatment effect at the presence of mediate variable. Our approach can also estimate the \emph{Direct Treatment Effect} where no mediate variable exists in observational data.
	This scenario implies the treatment $t$ is assumed to have a direct effect on the outcome $y$, i.e., $t\rightarrow y$. 
	In case the prior knowledge of $t\rightarrow y$ is known in practice, our approach can estimate \emph{Direct Treatment Effect} by merely removing \emph{mediate feature learning} component.
	Recall that \emph{debiasing confounder} adjusts the confounder variables by learning a treatment-invariant representation $\phi(\cdot)$, so that the treatment assignment is independent of the confounding bias.
% 	In addition, $\phi(\cdot)$ is regularized by the similarity loss~\eqref{eq:orth} to guarantee that the learned mediate features is treatment-specific.
	Without mediate variable $m$, $\phi(\cdot)$ is no longer regularized by the orthogonal constraint~\eqref{eq:orth} and becomes an unique cause of the outcomes. 
% 	Previous causal inference methods assume that all the variables are not affected by the intervention on treatment, which indicates that no mediate variable exists in the observational data.
% 	The \emph{mediate feature learning} component enables our approach to estimate the treatment effect at the presence of mediate variable.
    Then the learned $\phi(\cdot)$ is  directly feed into \emph{outcome prediction} for inferring treated and controlled outcomes, respectively.
    Finally, \emph{ITE} can be computed via Eq.~\eqref{eq:y}.

	%Define $h\in\mathcal{H}:\mathcal{R}\times\mathcal{R}\rightarrow\mathcal{Y}$ as a hypothesis function parameterized by $\Theta$ defined over the representation space of $\Phi$ and $\Psi$.

	%The  parameters  of  the  network  are  trained  to minimise the prediction loss $\mathcal{L}_{y}$ as
	%where $\lambda$ is the marginal probability of treatment $t=1$ and may vary for different instances $x_i$, $L(\cdot,\cdot)$ is the squared $L_2$-norm.

	%\section{Back-propagation with Optimal Transport Loss}
	%Weused the same base learning rate of0.0002, and used theADAM solver (Ba & Kingma, 2015) with momentum0.5.
	\section{Optimization}
	%\qli{gradient reversal layer: Domain-Adversarial Training of Neural Networks}
	%To obtain an update direction for $W_i$, $\Psi$ and $h$, we propose a general framework called DTANet for \emph{ITE} estimation with an end-to-end algorithm, regularized minimization procedure which simultaneously fits both a balanced representation of the data and a hypothesis for the outcome.Based on the this idea, we are seeking the parameters $(\boldsymbol{\Theta}_{\Phi},\boldsymbol{\Theta}_{t},\boldsymbol{\Theta}_{c},\boldsymbol{\Theta}_{h})$ and the optimal transport $T$ that deliver a saddle point of the final loss function ~\eqref{eq:obj}. 
	We consider the deep feed-forward network 
	that is trained to minimize the final loss function Eq.~\eqref{eq:obj} 
	using  mini-batch  stochastic  gradient  descent  with  the  Adam  optimizer~\cite{kingma2014adam}. 
	%in Tensorflow
	Specifically, 
	we propose an end-to-end algorithm that alternatively trains the parameters of the potential network, 
	the confounder network and the mediate feature  representation network with back-propagation.
	%The final loss function of our framework can be written as 
	\begin{equation}
	\begin{split}
	\mathcal{L}=\mathcal{L}_{y}+\lambda_1\mathcal{L}_{sim}+\lambda_2\mathcal{L}_{balan}
	\end{split}
	\label{eq:obj}
	\end{equation}
	where $\lambda_1$ and $\lambda_2$ are hyper-parameters that control the interaction of the loss terms during learning. 
	
	\subsection{Updating $\Psi_t$ and $\hat{y}_{t}$ }
	
	Based on Eq.~\eqref{eq:psi} and Eq.~\eqref{eq:y}, 
	the representation $\Psi_t$ and outcome $\hat{y}_{t}$ are parametrized by $V_t$ and $\Theta_t$, respectively. 
	%Outcome networks $\hat{y}_{t}(\cdot)$ and $\hat{y}_{c}(\cdot)$ are parameterized by $\Theta_t$ and $\Theta_c$ in~~\ref{eq:y}.
	%	The gradient of objective function~\eqref{eq:obj} including $\mathcal{L}_{y}$ and $\mathcal{L}_{sim}$. 
	Given the learning rate $\eta$, the gradients of objective function Eq.~\eqref{eq:obj} with respect to parameters $V_t$ and $\Theta_t$ are 
	%the parameterization of treated group is updated by
	\begin{equation}
	\nabla_{V_t}\mathcal{L}=\frac{\partial\mathcal{L}_{y}}{\partial \hat{y}_{t}}\frac{\partial \hat{y}_{t}}{\partial V_t}+\lambda_1 \frac{\partial\mathcal{L}_{sim}}{\partial V_t},\quad\nabla_{\Theta_t}\mathcal{L}= \frac{\partial\mathcal{L}_{y}}{\partial \Theta_t}
	\label{eq:gv}
	\end{equation}
	%The treated outcome hypothesis with $t=1$ can be similarly optimized by updating $\Theta^{k+1}_1$. 
	So the gradient descent updates the corresponding parameters of $\Psi_t$ and $\hat{y}_{t}$. 
	The update for $\Psi_c$ and $\hat{y}_{c}$ is similar to $\Psi_t$ and $\hat{y}_{t}$, 
	since they have similar optimization subproblems.

	\subsection{Updating $\Phi$}
	Recall that the confounding representation $\Phi$ in Eq.~\eqref{eq:phi} is parametrized by $W$. 
	Update the confounding representation $\Phi$ is non-trivial due to 
	the existence of optimal transport loss $\mathcal{L}_{balan}$ in Eq.~\eqref{eq:obj}. 
	The gradient of $\mathcal{L}$ w.r.t. the  $W$ is
	\begin{equation}
	\begin{split}
	\nabla_{W}\mathcal{L}= \nabla_{W}\mathcal{L}_{y}+\lambda_1\nabla_{W} \mathcal{L}_{sim}+\lambda_2\nabla_{W}\mathcal{L}_{balan}
	\end{split}
	\label{eq:gw}
	\end{equation}
	To compute the gradient of optimal transport loss $\mathcal{L}_{balan}$, 
	we regularize it by adding a strongly convex term 
	\begin{equation}
	\mathcal{R}(T)=-\frac{1}{\lambda_3}\sum_{i,j} T_{i,j} \log\boldsymbol{\gamma}_{i, j}
	\end{equation}
	that is the entropy ~\cite{benamou2015iterative} of $\boldsymbol{\gamma}$. 
	Then, we solve the regularized loss term by the Sinkhorn's iterations~\cite{cuturi2014fast}
	\begin{equation}
	\boldsymbol{\gamma}^{k}=\operatorname{diag}(\mathbf{u}^{k}) \mathbf{K}
	\operatorname{diag}(\mathbf{v}^{k})=\mathbf{u}^{k} \mathbf{1}_{n_t}^{\top} 
	\circ \mathbf{K} \circ \mathbf{1}_{n_c}( \mathbf{v}^{k})^{\top}
	\end{equation}
	where $\circ$ is element-wise multiplication, 
	the element $\mathbf{K}_{i,j}\allowbreak =\exp(-\lambda_3 \mathbf{C}_{i,j})$ in kernel matrix $K$ is computed based on $\mathbf{C}_{i,j}$ in Eq.~\eqref{eq:cm}, 
	and the updates of scaling vectors are 
	\begin{equation}
	\mathbf{v}^{k}=\frac{\mathbf{1}_{n_t} / n_t}{\mathbf{K}^{\top} \mathbf{u}^{k-1}}, 
	\quad \mathbf{u}^{k}=\frac{\mathbf{1}_{n_c} / n_c}{\mathbf{K} \mathbf{v}^{k}}
	\end{equation}
	Update the pairwise distance matrix between all treated and controlled pairs $C_{\Phi}$ with $W^{k-1}$ by Eq.~\eqref{eq:cm}.
	Then, we have
	\begin{equation}
	\nabla_{W}\mathcal{L}_{balan}=\frac{\partial \langle \boldsymbol{\gamma}^{k}, C_{\Phi}\rangle}{\partial W}
	\label{eq:gw3}
	\end{equation}
	%therefore the gradient of loss term $\mathcal{L}_{y}$ and $\mathcal{L}_{sim}$ w.r.t. their inputs is
	%\begin{equation}
	%    \begin{split}
	%        &\boldsymbol{\Theta}_y\leftarrow\boldsymbol{\Theta}_y-\mu\frac{\mathcal{L}_y}{}\\
	%        &W\leftarrow W-\nabla_{\Theta}\mathcal{L}_{y}(h_{\Theta}(x,t),y)\\
	%            &W\leftarrow W-\nabla_{W}\mathcal{L}_{y}(h(\Phi_W(x),t),y)\\
	%            &V_0\leftarrow 
	%            &V_1\leftarrow V_1-\nabla_{V_1}\mathcal{L}_{y}(h(\Phi(x),\Psi_V_1(x,t)),y)
	%    \end{split}
	%\end{equation}
	Apparently, 
	the gradients of $\nabla_{W}\mathcal{L}_{y}$ and $\nabla_W\mathcal{L}_{sim}$ are 
	\begin{equation}
	\begin{split}
	&\nabla_{W}\mathcal{L}_{y}=\lambda_t\frac{\partial\mathcal{L}_{y}}{\partial \hat{y}_t}\frac{\partial  \hat{y}_t}{\partial W}+\lambda_c\frac{\partial\mathcal{L}_{y}}{\partial \hat{y}_c}\frac{\partial  \hat{y}_c}{\partial W}\\
	&\nabla_{W}\mathcal{L}_{sim}=\frac{\partial\mathcal{L}_{sim}}{\partial W}
	\end{split}
	\label{eq:gw2}
	\end{equation}
	
	With all these computed gradients, 
	the steps of solving Eq.~\eqref{eq:obj} are shown in Alg.~\ref{alg:1}. 
	%We trained the model using the observed factual samples with $n_t$ treated samples $\mathcal{D}_t=(\boldsymbol{x}_{t_i},1,y_{t_i})_{i=1}^{n_t}$ and $n_c$ controlled samples $\mathcal{D}_c=(\boldsymbol{x}_{c_i},1,y_{c_i})_{i=1}^{n_c}$
	Note that the mediate feature  representation network and potential outcome network are trained only
	using the batch with the respective treatment, e.g., 
	the batch of treated individuals for treated features $\Psi_t(\cdot)$ and treated outcome $\hat{y}_t$.
	% mismatch between the training data and the DNN predictions is locally minimized.
	
	%Neural networks (NN) are a set of algorithms, inspired by the biological neural networks in brains, for classification and regression tasks. There are various types of NNs with different neuron connection forms and architectures, e.g., fully-connected neural networks (FC-NN), convolutional neural networks (CNN), and recurrent neural networks (RNN). 
	%Generally, training of a DNN is purely data-driven, and it consists of 
	
	%\qian{the intialization of ot, values for ot}
	\begin{algorithm}
		\begin{algorithmic}[1]
			\renewcommand{\algorithmicrequire}{\textbf{Input:}}
			\renewcommand{\algorithmicensure}{\textbf{Output:}}
			\caption{Treatment-Adaptive Network for Causal Inference} \label{alg:1}
			\Require Treated individuals $(\boldsymbol{x}_{t_i},y_{t_i})_{i=1}^{n_t}$ 
			and controlled individuals $(\boldsymbol{x}_{c_j},y_{c_j})_{j=1}^{n_c}$. 
			Adam hyper-parameters $\alpha$, $\beta_1$, $\beta_2$.
			scaling parameters $\lambda_0$, $\lambda_1$, $\lambda_2$, $\lambda_3$,  $\mathbf{u}=\boldsymbol{1}_{n_c}$
			%\State Train a linear regression function $f(x)=\boldsymbol{\beta}^{\top}\boldsymbol{x}^{CF}+\xi$ over the control samples $(x_i^{CF}, y_i^{CF})$
			%\State Construct $\Theta$, $\Phi$ and $\Psi$ using
			%\State $\boldsymbol{\nu}\leftarrow\frac{1}{n}\boldsymbol{1}_{n}$ 
			\While {not converged}
			\State Sample a treated batch $\mathcal{D}_{t}$ and controlled batch $\mathcal{D}_{c}$ %from observed data.
			\State Compute $\nabla_{W}\mathcal{L},\nabla_{V_t}\mathcal{L},\nabla_{V_c}\mathcal{L},\nabla_{\Theta_t}\mathcal{L}$, $\nabla_{\Theta_c}\mathcal{L}$
			\State Update $W,V_t,V_c,\Theta_t,\Theta_c$ by Adam optimizer
			%\State Compute $C(\Phi(x;W))$
			\State Compute representations $\Phi(\cdot;W),\Psi_t(\cdot;V_t),\Psi_c(\cdot;V_c)$
			%\State Compute potential outcomes $\hat{y}_t(;\Theta_t),\hat{y}_c(;\Theta_c)$
			%\State Compute the \emph{ITE} by
			\iffalse
			\State $W\leftarrow \text{Adam}(\nabla_{W}\mathcal{L},W,\mathcal{D}_{c_*},\mathcal{D}_{t_*})$
			\State $\{V_t\}\leftarrow \text{Adam}(\nabla_{V_t}\mathcal{L},V_t,\mathcal{D}_{t_*})$ 
			\State $\{V_t,V_c\}\leftarrow \text{Adam}(\nabla_{V_t}\mathcal{L},\mathcal{D}_{t_*})$ 
			\State $V_c\leftarrow \text{Adam}(\nabla_{V_c}\mathcal{L},\mathcal{D}_{c_*})$ 
			\State $\Theta_t\leftarrow \text{Adam}(\nabla_{\Theta_t}\mathcal{L},\mathcal{D}_{t_*})$ 
			\State $\Theta_c\leftarrow \text{Adam}(\nabla_{\Theta_c}\mathcal{L},\mathcal{D}_{c_*})$
			\fi
			%\State Compute the optimal matching $\hat{\boldsymbol{x}}_i^{F}$ for $\boldsymbol{x}_i^{F}$ using~~\ref{eq:bary_x}
			%\State Compute the counterfactual outcome $\hat{y}^{CF}$ by $f(\boldsymbol{x}^{CF})$
			\EndWhile
			\Ensure DTANet parameters $(W,V_t,V_c,\Theta_t,\Theta_c)$
			
		\end{algorithmic}
	\end{algorithm}

%		\gliMarker %TODO: GLi here
		
	\section{Experimental Results}
	
	Our deep model is a feed-forward neural network consisting of one confounder network, 
	two mediate feature  representation networks and two potential outcome networks. 
	Both the confounder network and the potential outcome network are 
	implemented as a three fully connected layers with 200 neurons. 
	The mediate feature representation network consists of 3 fully connected hidden layers.
	The activation function is the exponential linear unit (ELU). 
	The weights of all layers in each epoch is updated by the Adam optimizer with default settings. 
	We use the Adam optimiser with the initial learning rate of $\alpha=10^{-3}$, 
	decay rates $\beta_1=0.8$ and $\beta_2=0.95$. 
	Parameters $\lambda_0$ and $\lambda_3$ are empirically set to $0.5$ and $0.1$, 
	respectively. 
	We tune hyper parameters $\lambda_1,\lambda_2$ 
	via a grid search over combinations of 
	$\lambda_1\in[0.1,0.2], \lambda_2\in[0.3,0.45]$.

	%\qian{??}We additionally varied the weight of the imbalance penalty at random between 0.1, 1.0, and 10.0.

	\subsection{Datasets}
	
	\textbf{Real-world Data.} 
	We use real-world datasets, 
	i.e., \texttt{News}~\cite{johansson2016learning} and \texttt{JobsII}~\cite{vinokur1997mastery}.
	%~\footnote{\url{https://cran.r-project.org/web/packages/mediation/}}
	%Due to the space constraints, we defer descriptions on datasets and evaluation metrics to the appendix.
	\texttt{News} is a benchmark dataset designed 
	for counterfactual inference~\cite{johansson2016learning}, 
	which simulates the consumers' opinions on news items affected 
	by different exposures of viewing devices.
	%This motivates the assumption that consumer prefers to read certain News on mobile device. 
	This dataset randomly samples $n=5000$ news item 
	from NY Times corpus~\footnote{\url{https://archive.ics.uci.edu/ml/datasets/bag+of+words}}.
	Each sample is one new item represented by word counts 
	$\boldsymbol{x}_i\in\mathbb{R}^{d\times 1}$, 
	where $d=3477$ is the total number of words. 
	The factual outcome $y_i$ is the reader's opinion on 
	$\boldsymbol{x}_i$ under the treatment $t_i$. 
	The treatment represents two possible viewing devices, 
	where $t=0$ or $t=1$ indicates 
	whether the new sample is viewed via desktop and mobile $(t=1)$, 
	respectively. 
	The assignment of a news item $\boldsymbol{x}_i$ to a certain device $t$ is 
	biased towards the device preferred for that item.
	
	%IHDP dataset is a standard semi-synthetic causal inference benchmark collected from Infant Health and Development Program~\cite{hill2011bayesian}, where two-sides (factual and counterfactual) outcomes or treatment assignment are fully known.
	%IHDP studies the effect of the specialist visits and parent support on future cognitive and health status of infants. 
	%This dataset is based on a randomized experiment investigating the causal effect of home visits by specialists (treatment) on future cognitive scores (outcome). The dataset collects 747 individuals with 26 pre-treatment covariates measuring aspects of infants and their mothers, and 139 individuals are treated. 
	%The outcome is simulated IQ scores at age $3$. 
	%Briefly, the confounders correspond to collected measurements of the children and their mothers used during a randomized experiment that studied the effect of home visits by specialists on future cognitive test scores. The treatment assignment is then “de-randomized” by removing from the treated set children with non-white mothers; 
	%The treatment
	%an observational program designed to study the effect of the specialist visits and parent support on future cognitive and health status of infants. 
	
	\texttt{JobsII} dataset is collected from an observation study that 
	investigates the effect of a \emph{job training} (treatment) on the 
	outcome of one continuous variable of \emph{depressive symptoms}~\cite{vinokur1997mastery}. 
	Different from the treatment has direct causal effect on outcome in \texttt{News}, 
    the causal effect of the treatment on the outcome in \texttt{JobsII} is direct or indirect 
    via a mediate variable \emph{job-search self-efficacy}. 
    Because \emph{job-search self-efficacy} can be increased by \emph{job training} (treatment) and in turn affects the \emph{depressive symptoms} (outcome). 
    %We use the policy risk in~~\ref{eq:risk} as the metric for \texttt{JobsII}. 
    \texttt{JobsII} includes 899 individuals with 17 covariates, where 600 treated individuals with \emph{job training} and 299 controlled individuals without \emph{job training}. 
 %Each individuals have base-line c

	\textbf{Synthetic Data.}
	To illustrate our model could better handle both 
	hidden confounders and mediate variables, 
	we experiment on the simulated data of $n=1500$ samples 
	with $d$-dimensional covariates $(y,t,\boldsymbol{x},m)_{i=1}^{n}$. 
	For each $i$-th individual, 
	the dimension of the covariate $\boldsymbol{x}_i$ is set up to 100.
	To simulate the hidden confounding bias and noise, 
	we need to define several basis functions w.r.t. covariates $x$. 
	We follow the protocol used in~\cite{sun2015causal} and define ten basis functions as $f_{1}(x)=-2 \sin (2 x)$ $f_{2}(x)=x^{2}-1 / 3$, $f_{3}(x)=x-0.5$, $f_{4}(x)=e^{-x}-e^{-1}-1$, $f_{5}(x)=(x-0.5)^{2}+2$, $f_{6}(x)=\mathbb{I}_{\{x>0\}}$, $f_{7}(x)=e^{-x}$  $f_{8}(x)=\cos (x)$, $f_{9}(x)=x^{2}$, and $f_{10}(x)=x$.
	In addition to $\{g_1(x),\cdots,g_{10}(x)\}$, 
	we additionally define 5 basis functions for simulating mediate variable  influences 
	$g_{11}(x)=sin(x)-2*cos(5*x)$, $g_{12}(x)=-2*exp(x)$, 
	$g_{13}(x)=-2*x^2+1$, $g_{14}(x)=sin(3*x)$ and $g_{15}(x)=-2*cos(x/2)$.
	%li2017matching
	%$g_6(x)=e^{-x}$, $g_7(x)=x^2$, $g_8(x)=x$, $g_9(x)=\mathcal{I}_{x>0}$, $g_{10}(x)=cos(x)$, $g_{11}(x)=cos(x)$ and $g_{12}(x)=cos(x)$. 
	We also generate the binary treatment $t_i$ from a misspecified function 
	that if $\sum_{k=1}^5g_k(x)>0$ for $t_i=1$ and $t_i=0$ otherwise. 
	The mediate variable is  
	$m_i\sim\mathcal{N}(\sum_{k=1}^{5} g_{k+10}(x)+ct_i,1)$.
	%logistic function as $t_i\sim \text{Bernoulli}(1/(1+\exp{(-\sum_{i=1}^{d\cdot r}\delta\cdotx_i+\mathcal{N}(0,1)))})$ $r$ and $\delta$ indicates the ratio and strength of unobserved confounder on treatment. 
	The outcome is generated as follows. 
	\begin{equation}
	    y_i\sim \mathcal{N}\left(\sum_{k=1}^{5} g_{k+5}\left(\boldsymbol{x}_{k}\right)+at_i+b m_i, 1\right)
	\end{equation}
	The first five covariates are correlated to the treatment and the outcome, 
	simulating a confounding effect, 
	while the rest of them are noisy covariates.
	Following the routine of~\cite{rosenbaum1983central}, we use covariates $\{\boldsymbol{x}_1,\cdots,\boldsymbol{x}_5\}$ as informative variables that have confounding  effects  to  both  treatment and outcome. Causal inference works are all under the common simplifying assumption of ``no-hidden confounding'', i.e., all counfouders can be observed and measured from observed covariates. In other word, baseline methods can use covariates $\{\boldsymbol{x}_1,\cdots,\boldsymbol{x}_5\}$ as inputs to generate both treatment $t$ and outcome $y$ in the experiment.
	
	%The true causal effect (i.e., the ground truth value of ATT) in this dataset is 1.
	
	%We use a subset of the original data that 

	%Figure 1 shows overlaying model belief histograms for four  demographic  groups  and  their  barycenter  in  the Adult dataset. Wasserstein-1 Penalty effectively matches all group histograms to the barycenter after training for10,000 steps withβ= 100.

	%We consider two different estimation tasks.  One iswithin-sample,  where the task is to estimate \emph{ITE} for all individuals ina sample for which the (factual) outcome ofonetreatmentis observed.  This corresponds to the common scenario inwhich a cohort is selected once and not changed. This taskis non-trivial, as we never observe the \emph{ITE} for any individual. Theother isout-of-sample, where the goal is to estimate \emph{ITE} forindividuals withnoobserved outcomes.  This corresponds to theproblem of selecting the best treatment for anewpatient.Within-sample error is computed over both the training andvalidation sets, out-of-sample error over the test set
	\subsection{Baselines}
	
	We compare our method with the following four categories of baselines including
	(I) \emph{regression based methods}; 
% 	Ordinary Least Squares with treatment as a feature (OLS-1), 
% 	OLS with separate regressor for each treatment (OLS-2); 
	(II) \emph{classical causal methods}; 
% 	Doubly Robust Linear Regression (DR)~\cite{dudik2011doubly}, 
% 	Propensity Score Matching 
% 	 (PSM)~\cite{rosenbaum1983central}; 
	(III) \emph{tree and forest based methods}; 
% 	Bayesian Additive Regression Trees (BART)~\cite{hill2011bayesian}, 
% 	Causal Random Forest (CF)~\cite{wager2018estimation}; 
	(IV) \emph{representation based methods};
% 	Balancing Neural Network (BNN)~\cite{johansson2016learning}, 
% 	Treatment Agnostic Representation Network (TARNet)~\cite{shalit2017estimating} 
% 	and Counterfactual Regression Network (CFRNet)~\cite{shalit2017estimating}. 
	%All benchmarks are implemented in Python, with the exception of DR being implemented in R\footnote{https://github.com/gregridgeway/fastDR}. 
	%implementation of BayesTree R-package~\footnote{https://cran. r-project.org/package=BayesTree/}
	%We used the R libraries bartMachine, grf, and DR for the implementation of BART, causal forests and DR, respectively.

	\begin{itemize}
	\item 
	 OLS-1~\cite{goldberger1964econometric} (I): this method takes the treatment as an input feature and 
	predicts the outcome by least square regression.
	%predicts the factual outcome by $t=1$ and control outcome by $t=0$.
	\item
	OLS-2~\cite{goldberger1964econometric} (I) : this  
	uses two separate least squares regressions 
	to fit the treated and controlled outcome respectively.  
	%Causal effect is estimated by the difference between the two outcomes.
	\item  TARNet~\cite{shalit2017estimating} (I): this method is \emph{Treatment-Agnostic Representation Network} that captures non-linear relationships underlying features to fit the treated and controlled outcome.
	\item 
	 PSM~\cite{rosenbaum1983central} (II): this method refers to \emph{Propensity Score Matching} that matches the controlled individuals 
	which received no treatment 
	with those treated individuals which received the treatment,
	based on the absolute difference between their propensity scores.

	\item 
	 DR~\cite{dudik2011doubly} (II): this method refers to \emph{Doubly Robust Linear Regression} which is a combination 
	of regression model and propensity score estimation model 
	to estimate the treatment effect robustly.
	%misspecification may occur if any influential covariates are not included in the propensity score estimation model or if there are any misspecified forms of covariates, such as interaction effect, higher-order terms, or non-linear trends. Model misspecification can happen not only for the propensity score estimation model, but also in the outcome regression model. In this situation, using doubly robust methods will increase the accuracy of outcome estimation after propensity score adjustments. Doubly robust estimation incorporates outcome regression model and propensity score model in treatment effect estimation, which is robust to one model misspecification (either regression model or propensity score model). Doubly robust procedures were found to reduce more bias than just using one propensity score procedure alone (Shadish et al. 2008). Therefore, doubly robust estimation is increasingly used when implementing propensity score methods.

	\item  BART~\cite{hill2011bayesian} (III): this method is \emph{Bayesian Additive Regression Trees} that directly applies a prior function 
	on the covariate and treatment to estimate the potential outcomes, 
	i.e., 
	Bayesian form of the boosted regression trees.
	%BART can avoid the overfitting by the prior function specification. 
% 	The number of regression trees is set as $200$.
	
	\item  CF~\cite{wager2018estimation} (III): this method refers to \emph{Causal Forest} as an extension 
	of random forest. It includes a number of causal trees and estimates the treatment effect on the leaves.

	\item BNN~\cite{johansson2016learning} (IV): this is called \emph{Balancing Neural Network} that attempts to learn a balanced representation by minimizing the similarity 
	between the treated and the controlled individuals 
	for counterfactual outcome prediction.
% 	As suggested in~\cite{johansson2016learning},
% 	we use 2-ReLU representation-only layers, 
% 	2-ReLU layers for the added treatment variable, 
% 	and a single linear output layer.

	\item CFRNet~\cite{shalit2017estimating} (IV): this method refers to \emph{Counterfactual Regression Networks} that attempts to find balanced representations by minimizing the Wasserstein distance between the treated and controlled individuals.

\end{itemize}

	For hyper-parameters optimization, 
	we use the default prior or network configurations 
	for TARNet~\cite{johansson2016learning}, BART~\cite{hill2011bayesian}, CFRNet~\cite{shalit2017estimating}, BNN~\cite{johansson2016learning}. 
	%PSM was implemented as described in~\cite{hill2011bayesian}. 
	For PSM, 
	we apply 5-nearest neighbour matching with replacement, 
	and impose a nearness criterion, i.e., caliper=0.05. 
	The number of regression trees in BART is set to 200, 
	and CF consists of 100 causal trees.
	Parameters in other benchmarks are tuned to achieve their best performances. All datasets for all models are split as training/test sets with a proportion of 80/20, and 20\% of the training set are validation set. The within-sample error is calculated over  validation sets, and out-of-sample error is calculated over test set.

	%The resolution parameter of the persistence image is selected from and the kernel parameter is selected from {0.01, 0.05, 0.1, 0.15, 0.2}, resulting in 6 ⇥ 5 = 30 pairs of parameter settings
	%We chose hyperparameters at random from predefined ranges
	%For the IHDP and sythetic datasets we respectively used 30 and 10 optimisation runs for each method using randomly selected hyperparameters from predefined ranges.
	
	%for a maximum of 400 with an early stopping patience of 30\qian{values...}. 

	\subsection{Metrics}
	
	%\subsubsection{Treatment Effect}
	The goal of causal inference is to estimate the treatment effect 
	at the individual and population level. 
	Previous causal effect estimation algorithms are prominently evaluated 
	in terms of both levels.
% 	Thus on the treated individuals, 
% 	we can have three metrics including the individual treatment effect (\emph{ITE}), 
% 	 average treatment effect (ATE) and  average treatment effect (ATT). 
	For the individual-based measure $\tau_{ITE}$ defined in Eq.~\eqref{eq:ite}, 
	we have \emph{Precision in Estimation of Heterogeneous Effect (PEHE)}~\cite{hill2011bayesian}
	\begin{equation}
	\epsilon_{\mathrm{PEHE}}=\frac{1}{n} \sum_{i=1}^{n} \left(\tau_{ITE}(i)-\hat{\tau}_{ITE}(i)\right)^2  
	\end{equation} 
	where $\hat{\tau}_{ITE}(i)$ is the estimated individual treatment effect by $\hat{y}_i(1)-\hat{y}_i(0)$.
	%$\epsilon_{\mathrm{PEHE}}=\frac{1}{n} \sum_{i=1}^{n} \left(\hat{y}_i(1)-\hat{y}_i(0)-(y_i(1)-y_i(0))\right)^2$ 
	%where $\hat{y}_i(1)-\hat{y}_i(0)$ is the predicted individual effect.
	%One simulated data and \texttt{News} dataset, 
	For the population level,
	we use mean absolute error to evaluate models. 
	For instance, given the ground truth $\tau_{ATE}$ and the inferred
   $\hat{\tau}_{ATE}$ in Eq.~\eqref{eq:ite2}, the mean absolute error on \emph{ATE} is
   \begin{equation}
	\begin{split}
	    \epsilon_{ATE}=|\hat{\tau}_{ATE}-\tau_{ATE}|
	\end{split}
	\end{equation}
	Similarly, the mean absolute error to evaluate performance at population level is defined as follows:
	\begin{equation}
	\begin{split}
	    &\epsilon_{ATT}= |\hat{\tau}_{ATT}-\tau_{ATT}|\\
	    &\epsilon_{MTE}=|\hat{\tau}_{MTE}-\tau_{MTE}|,\quad\epsilon_{DTE}= |\hat{\tau}_{DTE}-\tau_{DTE}|
	\end{split}
	\end{equation}

	\begin{table*}[!hbt] 
		\centering
		\caption{In-sample evaluation on \texttt{News} and \texttt{JobsII}.}
		\begin{tabular}{cccccc}
			%\begin{tabular}{cccccc}
			\toprule
			\multirow{2}{*}{Method} &    \multicolumn{3}{c}{\texttt{News}}&\multicolumn{2}{c}{\texttt{JobsII}}
			\\ \cmidrule{2-6}
			& $\sqrt{\epsilon_{\mathrm{PEHE}}}$      &      $\epsilon_{\mathrm{ATE}}$    &   $\epsilon_{\mathrm{ATT}}$ &      $\mathcal{R}_{pol}$          &         $\hat{\epsilon}_{\mathrm{ATT}}$      \\
			\midrule
			
			OLS-1     &  5.2 $\pm$ 0.1       & 0.90 $\pm$ 0.3 & 0.89 $\pm$ 0.2 &   2.30 $\pm$ 0.2 & 0.02 $\pm$ 0.0 \\
			OLS-2 & 3.5 $\pm$ 0.2     &  0.45 $\pm$ 0.0 &   0.64 $\pm$ 0.1 &   2.37 $\pm$ 0.6 & 0.02 $\pm$ 0.0\\
			PSM &  4.8 $\pm$ 1.0 &    2.72 $\pm$ 0.7 &    2.62 $\pm$ 1.0 & 2.67 $\pm$ 0.5 & 0.02 $\pm$ 0.0 \\
			DR  & 4.7 $\pm$ 0.1     & 2.57 $\pm$ 0.6& 1.42 $\pm$ 0.2   &  2.41 $\pm$ 0.7 & 0.02 $\pm$ 0.0 \\
			BART  & 5.2 $\pm$ 0.1    &  1.57 $\pm$ 0.5  & 1.05 $\pm$ 0.8 &  1.94 $\pm$ 0.4   & 0.05 $\pm$ 0.0 \\
			CF  &  4.6 $\pm$ 0.2  & 1.62 $\pm$ 0.1  & 2.19 $\pm$ 1.3   & 1.79 $\pm$ 0.2   & 0.06 $\pm$ 0.0 \\
			%BNN& 2.21(2.2) $\pm$ 0.115(0.1)  & 0.37 $\pm$  0.03 &     $\pm$       & $\pm$ \\ 
			BNN& 4.8 $\pm$ 0.2       & 0.65 $\pm$ 0.0 & 0.97 $\pm$ 0.0  &   1.78 $\pm$ 0.1     & 0.05 $\pm$ 0.0  \\ 
			TARNet& 1.3 $\pm$ 0.2 &0.28 $\pm$ 0.0  & 0.28 $\pm$ 0.0  &1.67 $\pm$ 0.2     & 0.04 $\pm$ 0.0 \\
			CFRNet& 0.8 $\pm$ 0.3 & 0.26 $\pm$ 0.0 & 0.24 $\pm$ 0.0  &   1.55 $\pm$ 0.5     & 0.04 $\pm$ 0.0\\ 
			DTANet &  \textbf{0.6} $\pm$ 0.3 & \textbf{0.25} $\pm$ \textbf{0.0} &   \textbf{0.21} $\pm$ \textbf{0.0}& \textbf{1.40} $\pm$ 0.6     & \textbf{0.01} $\pm$ \textbf{0.0} \\ \bottomrule
			%\multicolumn{5}{c}{shape classification for \texttt{SHREC2014}}\\
			%\hline
			%SYN & 97.6  $\pm$  3.1 & 96.3  $\pm$  2.8 &     97.1  $\pm$  2.2      & \textbf{98.3}  $\pm$  0.8 \\
			%REAL & 62.5  $\pm$  2.2 & 59.5  $\pm$  2.3 &     64.1  $\pm$  2.4      & \textbf{67.2}  $\pm$  0.7 \\\hline
		\end{tabular}
		\label{tb:rs}
	\end{table*} 
	\begin{table*}[!hbt]
		\centering
		\caption{Comparison results on the simulated dataset.}
		\begin{tabular}{ccccccccc}
			\toprule
			\multirow{2}{*}{Method} &    \multicolumn{4}{c}{In-sample}&\multicolumn{4}{c}{Out-of-sample}
			\\ \cmidrule{2-9}
			& $\sqrt{\epsilon_{\mathrm{PEHE}}}$      &      $\epsilon_{\mathrm{ATE}}$    &   $\epsilon_{\mathrm{ATT}}$ &   $\epsilon_{\mathrm{MTE}}$ & $\sqrt{\epsilon_{\mathrm{PEHE}}}$      &      $\epsilon_{\mathrm{ATE}}$    &   $\epsilon_{\mathrm{ATT}}$&$\epsilon_{\mathrm{MTE}}$   \\
			\midrule
			OLS-1  &  5.43 $\pm$ 0.3 & 3.07 $\pm$ 0.4 &  3.06 $\pm$ 0.5& 2.15 $\pm$ 0.4 &  6.06 $\pm$ 0.5&  3.11 $\pm$ 0.4  &3.09 $\pm$ 0.6 &2.28 $\pm$ 0.4\\
			OLS-2 &3.24 $\pm$ 0.4  & 2.43 $\pm$ 0.2  &2.45 $\pm$ 0.5 & 1.53 $\pm$ 0.6    &4.92 $\pm$ 0.5     & 3.03 $\pm$ 0.6& 2.73 $\pm$ 0.6& 2.01 $\pm$ 0.5 \\
			PSM &5.00 $\pm$ 0.3  & 3.21 $\pm$ 0.2  &2.56 $\pm$ 0.5& 1.63 $\pm$ 0.4 &7.91 $\pm$ 0.5     & 4.06 $\pm$ 0.6& 2.33 $\pm$ 0.5& 1.39 $\pm$ 0.3
			\\
			DR  &4.50 $\pm$ 0.1  & 3.40 $\pm$ 0.2  &2.71 $\pm$ 0.5& 1.78 $\pm$ 0.5&6.91 $\pm$ 0.2     & 4.10 $\pm$ 0.1& 4.40 $\pm$ 0.2& 3.57 $\pm$ 0.3  \\
			BART & 3.10 $\pm$ 0.2 &   2.70 $\pm$ 0.1 & 2.90 $\pm$ 0.1& 1.85 $\pm$ 0.3 &   3.80 $\pm$ 0.3  & 3.01 $\pm$ 0.2& 2.98 $\pm$ 0.1& 1.95 $\pm$ 0.2 \\
			CF  & 1.95 $\pm$ 0.2& 1.21 $\pm$ 0.4 &  1.25 $\pm$ 0.2&  1.02 $\pm$ 0.2 &  2.63 $\pm$ 0.4 &  2.32 $\pm$ 0.2 & 1.33 $\pm$ 0.3& 1.41 $\pm$ 0.4\\ 
			%BNN& 2.21(2.2) $\pm$ .115(0.1)  & 0.37 $\pm$  0.03 &     $\pm$       & $\pm$ \\ 
			BNN &1.69 $\pm$ 0.4  & 1.20 $\pm$ 0.3  & 1.20 $\pm$ 0.1&0.78 $\pm$ 0.2 & 2.51 $\pm$ 0.3   & 2.42 $\pm$ 0.2& 2.05 $\pm$ 0.4 & 1.32 $\pm$ 0.5  \\
			TARNet &1.05 $\pm$ 0.2  & 0.82 $\pm$ 0.1&0.43 $\pm$ 0.1&0.35 $\pm$ 0.1 & 1.77 $\pm$ 0.2   & \textbf{0.73} $\pm$ \textbf{0.0}    &0.77 $\pm$ 0.1&0.45 $\pm$ 0.2
			\\ 
			CFRNet&  1.04 $\pm$ 0.2 &0.69 $\pm$ 0.1& 0.45 $\pm$ 0.2 &0.32 $\pm$ 0.1 &  1.62 $\pm$ 0.3  &0.87 $\pm$ 0.2& 0.66 $\pm$ 0.1& 0.34 $\pm$ 0.1 \\ 
			DTANet &  \textbf{0.86 }$\pm$ \textbf{0.1}  &\textbf{0.57} $\pm$ \textbf{0.1}& \textbf{0.34} $\pm$ 0.4& \textbf{0.27} $\pm$ 0.3&\textbf{1.37} $\pm$ 0.4  &0.85 $\pm$ 0.4  &\textbf{0.54} $\pm$ 0.1 &\textbf{0.32} $\pm$ 0.1 
			\\ \bottomrule
			%\multicolumn{5}{c}{shape classification for \texttt{SHREC2014}}\\
			%\hline
			%SYN & 97.6 $\pm$ 3.1 & 96.3 $\pm$ 2.8 &     97.1 $\pm$ 2.2      & \textbf{98.3} $\pm$ 0.8 \\
			%REAL & 62.5 $\pm$ 2.2 & 59.5 $\pm$ 2.3 &     64.1 $\pm$ 2.4      & \textbf{67.2} $\pm$ 0.7 \\\hline
		\end{tabular}
		\label{tb:simulated}
		
	\end{table*}

	The above metrics can not be applied on  \texttt{JobsII}, because there is no ground truth for \emph{ITE} in \texttt{JobsII}. Specifically, \texttt{JobsII} doesn't include two potential outcomes for an individual under both treated and controlled condition.
	Instead, in order to evaluate the quality of \emph{ITE} estimation,
	the policy risk is used as the metric on \texttt{JobsII} dataset.
	The policy risk $\mathcal{R}_{pol}$~\cite{shalit2017estimating} is used as the metric to 
	measure the expected loss if the treatment is taken according to \emph{ITE} estimation. 
	\begin{equation}
	\begin{split}
	\mathcal{R}_{p o l}(\pi_f)=1-\mathbb{E}\left[\hat{y}_t \mid  \pi_f=1\right] p(\pi_f=1)\\
	-\mathbb{E}\left[\hat{y}_c \mid  \pi_f=0\right] p(\pi_f=0) 
	\label{eq:risk}
	\end{split}
	\end{equation}
	In our case, 
	we let the policy be to treat,  
	$\pi_f=1$ if $\hat{y}_t-\hat{y}_c>0$, 
	and to not treat, $\pi_f = 0$, otherwise. 
	We divide benchmark data into a training set (80\%) and an out-of-sample testing set (20\%), 
	and then evaluate those three metrics on the testing sample in 100 different experiments.
	For all the metrics, 
	the smaller value indicates the better performance.  

	\subsection{Results and discussion}

	\subsubsection{Treatment effect estimation.}
	We first compare all methods on the task of treatment effect estimation. We perform this task on two real-world datasets (i.e., \texttt{News} and \texttt{JobsII}) and one synthetic dataset with binary treatment.
	The performance of all methods on \texttt{News} and \texttt{JobsII}  are shown in Table~\ref{tb:rs}.
	The results for \texttt{News} and \texttt{JobsII} are reported by employing in-sample evaluation.
	In-sample evaluation refers to evaluate the treatment effect of the common scenario where one potential outcome under treatment variable $t=1$ or $t=0$ is observed for each individual~\cite{shalit2017estimating}. For example, a patient has received a treatment and is observed with the health outcome.
	The error of in-sample evaluation is computed over validation set.
	%In-sample case evaluates the performance metric on the training dataset.
	%In the within-sample case, the performance metric is measured on the training dataset, and the out-of-sample case is on the test dataset. 
	Apparently, our DTANet performs the best on \texttt{News} dataset.
	The representation methods perform better than other baselines for \texttt{News} in all metrics.
	This is mainly because they reduce the confounder bias by balancing the covariates between treated and controlled individuals. 
	%CF and BART outperforms slightly better than BNN on $\epsilon_{ATE}$ and $\epsilon_{ATT}$. This might be because BNN has more parameters to be optimized and \texttt{News} is a relatively small dataset. 

	One major contribution of our DTANet is to 
	alleviate the bias of treatment effect estimation due to the ignorance of mediate variables. 
	Different from \texttt{News}, \texttt{JobsII} involves the mediate variable $m$ referring to the level of workers' job search self-efficacy. 
	The outcome is a measure of depression for each worker. 
	Compared with the results of \texttt{News}, 
	the performance of the representation learning is degraded, 
	i.e., the worst $\epsilon_{\mathrm{ATT}}$.
	The comparison baselines neglect the mediate-specific information introduced by the mediate variables.
	This verifies that 
	neglecting the mediate variable leads to the unstable estimation of treatment effect.
	Our method has both balancing property and treatment-adaptive ability to 
	improve the accuracy of treatment effect estimation, 
	which brings the best performance to both datasets.

	To further evaluate the generalization of baseline methods, 
	we perform the out-of-sample evaluation on the synthetic dataset to estimate ITE for individuals with no observed potential outcome. This refers to the scenario where a new patient arrives and the goal is to choose the best possible treatment. The error of out-of-sample is computed over the test set.
	The out-of-sample aims to estimate ITE for units with no observed outcomes. This corresponds to the case where a new patient arrives without taking any treatment and the goal is to select better treatment between treatment A and B. The within-sampling setting refers to the case where a patient has already taken treatment A but we then want to select the better treatment between A and a new treatment B. In-sample error is computed over the validation sets, and out-of-sample error over the test set.
	Table~\ref{tb:simulated} is obtained 
	by setting $a=2$, $b=0.5$ and $c=1$ for the synthetic data. 
	Their performance is worse than our DTANet on the simulated data. 
	%Similar to the performance on real-world dataset, our  proposed  method  consistently  performs  best under  all three  evaluation  metrics among the contending methods.
	This observation verifies that 
	DTANet uses mediate feature  representation for the unmeasured mediate variables 
	and thus can improve treatment effect estimation.
	The out-of-sample setting is much more challenging than the in-sampling setting. Our approach produces a confounding representation that is invariant for both treatments via orthogonal projection constraint. This guarantees the inputs of confounding representation are uncontaminated with information unique to each treatment. Consequently, the potential outcome predictor trained on confounding representation is better able to generalize across different treatments, and further to provide a basis for the estimation of unbiased treatment effect.
	%Compared with other representation based methods, the proposed DTANet not only considers , but also preserves the local treatment-specific and informative information in the original feature space.
	%BNN obtains better $\epsilon_{PEHE}$ than CF and BART on both dataset as it considers the balanced property across treatment and control groups. 

	%While TARNet doesn't have any regularization in the representation space, and its outcome prediction network is dichotomous. 
	
	\begin{figure}[!htb]
    \centering
    \includegraphics[width=3.2in]{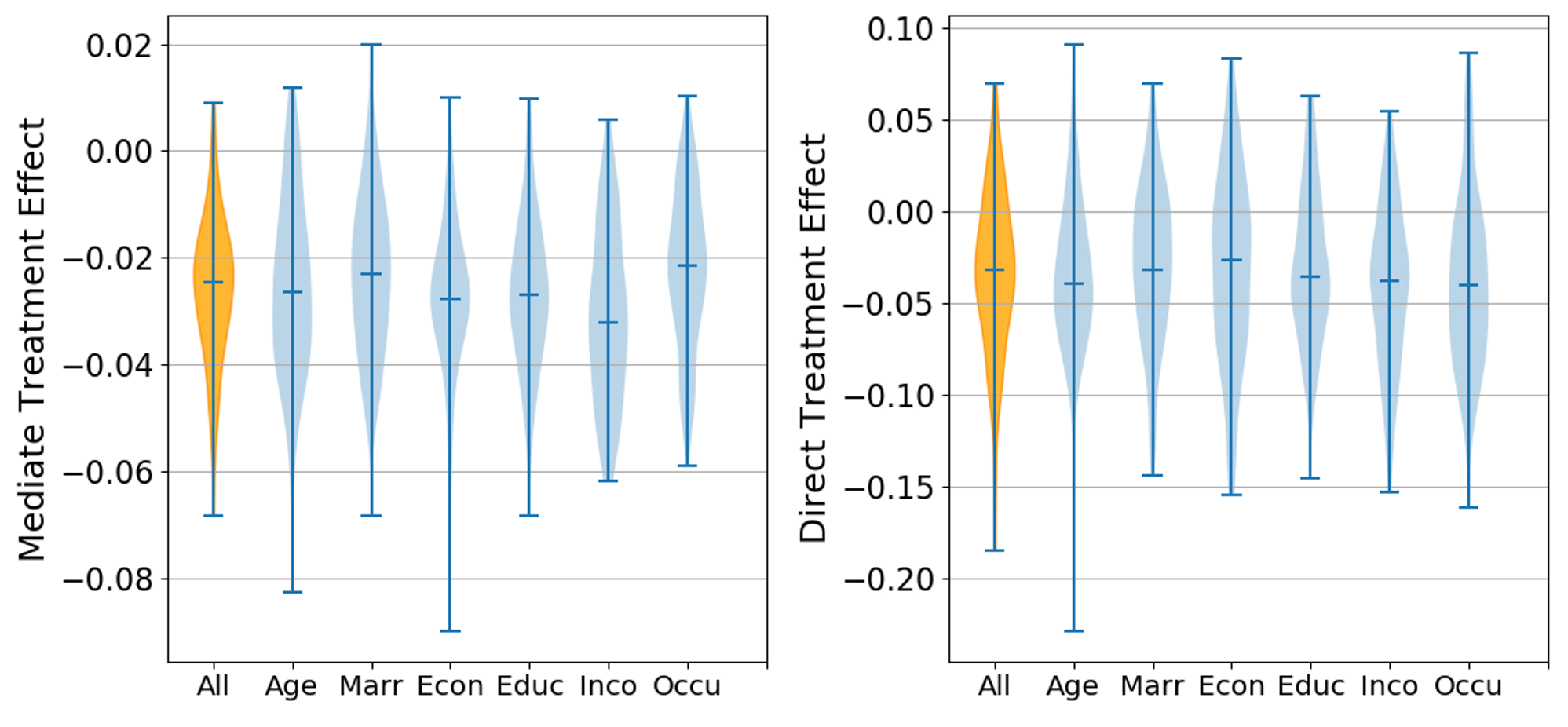}
    \caption{Our DTANet results on \texttt{JobsII}: The distributions of estimated treatment effect caused by different covariates for our DTANet.}
    \label{fig:contri}
\end{figure}

	\subsubsection{Causal Explanations}
	The covariate/feature importance for the predictions 
	is a simple but effective solution for explanations. 
	Since our DTANet is causality-oriented, 
	this experiment attempts to provide causal explanations 
	for the estimated treatment effect by analyzing the contributions of input covariates.

	\begin{table}[]
	\centering
	\caption{The distance (unit is $10^{-3}$) between the distribution of \emph{Mediate Treatment Effect} / \emph{Direct Treatment Effect} (using entire covariates) and that of excluding particular covariate.}
    \begin{tabular}{ccccccc}
    \midrule
    & Age  & Marr & Econ & Educ & Inco & Occu \\\hline
    Mediate & 3.98 &  4.09 & 5.75& 1.97 & 2.09 & 3.12 \\
    Direct &  10.4& 9.03& 9.98 & 5.62 & 7.13 &6.11  \\
    \bottomrule
\end{tabular}
	\label{tb:ws}
\end{table}
	%This experiment confirms that the existence of mediate variables that are affected by the observed covariates with different degrees.
	%One the other hand, this experiment can further exploit the contribution of each covariate to the treatment effect estimation 
	
	%In addition to producing accurate estimates of covariates importance, we propose the use of bootstrap ensemble methods, specifically using bootstrap resampling
	%we conduct two Monte Carlo simulations each with two different sample sizes (100 and 500) for a total of \texttt{JobsII} dataset. 
	
	%We reassigned outcomes and treatments with a new random seed for each repetition.
	%Figure ~\ref{fig:contri} reports the importance of covariates on the treatment effect. y-axis is Mediate Treatment Effect. Each element in x-axis is the specific covariate excluded from entire covariates. Because we ran DTANet on JobII 100 trials, we can get their distributions. 
	
	To accurately quantify the covariates importance, 
	we repeatedly run our DTANet on \texttt{JobsII} and predict the treatment effect with different input covariates. 
	We run DTANet on \texttt{JobsII} 100 trials, so we get 100 results and then obtain their distributions.
	As shown in Figure~\ref{fig:contri}, $y$-axis is \emph{Mediate/Direct Treatment Effect} and $x$-axis is the specific covariate excluded from entire covariates.
	The batch results colored in orange are gained by inputting all covariates. Each batch in blue corresponds to the estimated treatment effect by DTANet without a specific covariate. The estimated \emph{Mediate Treatment Effect} is significantly different from zero, suggesting that treatment (\emph{job training}) changes the mediate variable (\emph{job-search self-efficacy}), which in turn changes the outcome (\emph{depressive symptoms}). 
	We find that three covariants,
	\emph{Econ} (economic hardship), 
	\emph{Marr} (marital status) and \emph{Age}, 
	are the main causes of the treatment effect, 
	which is consistent with study~\cite{vinokur1997mastery}.
	Particularly, we consider the distribution of \emph{Mediate/Direct Treatment Effect} produced by entire covariates as the baselines.
	As shown in Figure~\ref{fig:contri}, 
	the distributions of excluding \emph{Econ}, 
	\emph{Marr} and \emph{Age}, respectively, 
	are the three most significant ones
	that extend the baseline distribution with larger ranges. To further quantify the differences between baseline distributions and the distributions of excluding covariates, we resort to the original Wasserstein distance~\cite{peyre2019computational} as a metric in Table~\ref{tb:ws}.
	Particularly, we use the function \texttt{wasserstein\_distance} in python library \texttt{SciPy}\footnote{https://docs.scipy.org/doc/scipy/reference/generated/scipy.stats.wasser\\stein\_distance.html} to compute the Wasserstein distance between two distributions. For example, $3.98\times 10^{-3}$ is the Wasserstein distance between the distribution of \emph{Mediate Treatment Effect} with entire covariates and the distribution excluding covariate \emph{Age}.
	According to the results in Table~\ref{tb:ws}, the distributions of \emph{Econ}, 
	\emph{Marr} and \emph{Age} have larger Wasserstein distances from the baseline distributions. In other words, these three covariates can significantly impact the \emph{Mediate/Direct Treatment Effect}.
	This conclusion validates that the mediate feature representation in our DTANet method can generate effective causal explanations for the \emph{Mediate Treatment Effect} estimation. On the other hand, the covariates contribute similar amounts to \emph{Direct Treatment Effect} except \emph{Age}. We can deduce that \emph{Age} is the common cause for the treatment (\emph{job training}) and outcome (\emph{depressive symptoms}), i.e., the confounder.

\begin{figure}[!htb]
    \centering
    \begin{subfigure}{.48\textwidth}
      \centering
      \includegraphics[width=1\linewidth]{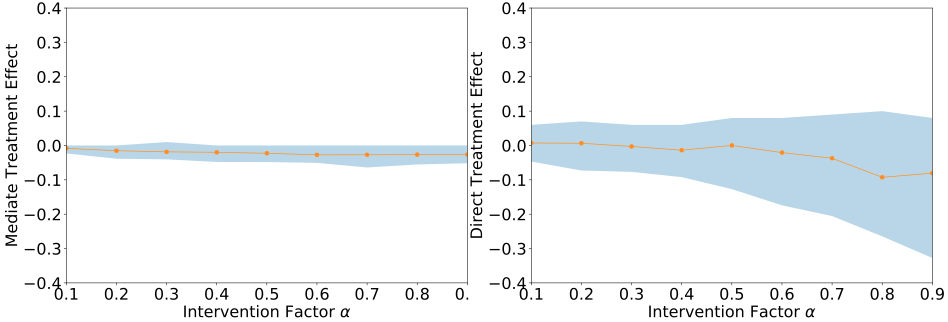}
    \end{subfigure}%
    % \\
    % \begin{subfigure}{.4\textwidth}
    %   \centering
    %   \includegraphics[width=.95\linewidth]{fig/quantile_right.pdf}
    % \end{subfigure}
    %\includegraphics[width=3.2in,height=1.6in]{fig/quantile.png}
    \caption{Our DTANet results on \texttt{JobsII}: the comparison of changes in estimated treatment effects caused by doing an intervention on the mediate variable. The blue cover represents 95\% confidence interval of the change.}
    \label{fig:quanti}
\end{figure}

	Figure~\ref{fig:quanti} demonstrates the estimated treatment effect when intervening on the mediator \emph{job search self-efficacy}. The left figure shows magnitude of the estimated \emph{Mediate Treatment Effect} increases slightly as one moves from lower to higher intervention factor. But the change is small, indicating the \emph{Mediate Treatment Effect} is relatively constant across the distribution. In contrast, the estimated direct effects vary substantially across different intervention factors, although the confidence intervals are wide and always include zero.

	 \subsubsection{Robustness analysis} 
	 There may exist unobserved confounders that causally affect both the mediator and outcome even after conditioning on the observed treatment and pre-treatment covariates. 
	Therefore, 
	we investigate the robustness of our DTANet to unmeasured confounding factor $\rho$. 
	The robustness analysis is conducted by varying the value of $\rho$ and examining how the estimated treatment effect changes.
	We define $\rho$ as the correlation between the error terms in the mediator and the outcome models.  
	This is reasonable, since unobserved confounder can bias 
	both estimation of  mediator and  outcome, 
	which further leads to unexplained variance or errors. 
	If unobserved confounder affects mediator and outcome, 
	we expect $\rho$ is non-zero.
	\begin{figure}[!htb]
    \centering
    % \begin{subfigure}{.4\textwidth}
    %   \centering
    %   \includegraphics[width=.95\linewidth]{fig/sens_left.pdf}
    % \end{subfigure}%
    % \\
    \begin{subfigure}{.48\textwidth}
      \centering
      \includegraphics[width=1\linewidth]{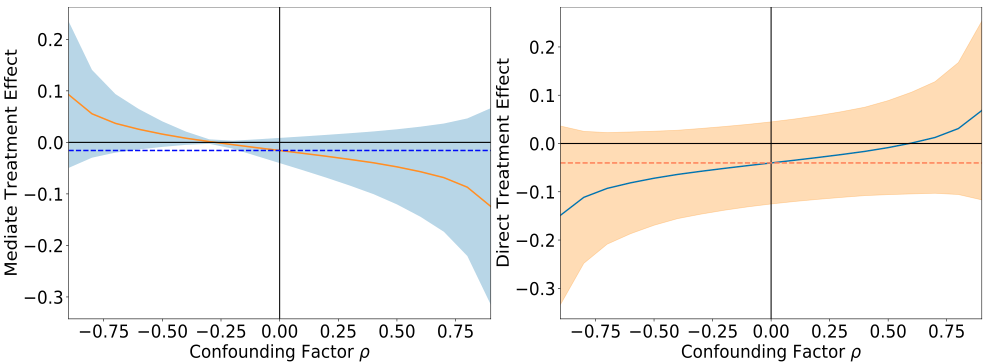}
    \end{subfigure}
    \caption{Robustness analysis of our DTANet on unobserved confounder. The dashed line represents the estimated mediation treatment effect. The areas represent 95\% confidence interval for \emph{Mediate Treatment Effect} at each $\rho$. The solid line represents the estimated average mediation effect at different values of $\rho$.}
    \label{fig:sen}
\end{figure}
	 
	 The estimates with potential outcome framework in Section~\ref{sec:pof} are identified if the ignorability assumption holds. However, it is possible that this assumption doesn't holds in practice. Thus, we next ask how sensitive these estimates are to violations of this assumption using our method. Figure~\ref{fig:sen} shows the estimated mediator treatment effect 
	and \emph{Direct Treatment Effect} against different values of $\rho$, where $y$-axis is the treatment effect and $x$-axis is the confounding factor.
	The true \emph{Mediate Treatment Effect} and \emph{Direct Treatment Effect} marked 
	as dash horizontal lines are -0.16 and -0.04, respectively. 
	That means no unobserved confounders exists for mediator and outcomes (i.e., $\rho=0$).
	The left figure shows the confidence intervals for \emph{Mediate Treatment Effect} (i.e., treatment effect due to mediation variable) covers the value of zero only under $\rho=-0.3$.
	 The \emph{Mediate Treatment Effect} 
	 is statistically indistinguishable from zero at the 95\% level when the parameter $\rho<-0.3$.
	Potentially, parameter $\rho$ should be higher than $0.3$ so that the effect will be insignificant in the left figure, however such low $\rho$ value is unlikely to happen in practice. In other words, treatment effect estimation by our DTANet is robust to 
	possible unobserved confounders in varying degrees.
	%This means that our DTANet would take a low probability chance to mis-estimates the treatment effect.

	%For the counterfactual task, we see that for small proxy noise all methods perform similarly.
	%A machine trained in one environment cannot be expected to perform well when environmental conditions change, unless the changes are localized and identified.

	\section{Conclusion}
	
	Individual treatment effect (ITE) estimation is one major goal of causal inference, 
	which aims to reduce the treatment assignment bias caused by the confounders.  
	Although recent representation based methods achieve satisfactory computational accuracy, 
	they overlook the unique characteristics of the treatment under different $do$ interventions. 
	Moreover, 
	the confounding representation from original covariates
	is easily affected by the treatment, 
	which violates the fact that confounder is irrelevant to treatment after $do$ intervention.
	In order to overcome above challenges in individual treatment estimation (ITE), 
	we propose an end-to-end model DTANet 
	to learn the confounding representation by optimal transport, 
	and it satisfies the treatment-invariant property introduced by doing an intervention. 
	Meanwhile, 
	by the proposed orthogonal projection strategy, 
	DTANet is capable of capturing the mediate features that are treatment-specific and
	are informative for the outcome prediction. 
	The effectiveness of DTANet is verified by both empirical and theoretical results.	
\bibliographystyle{spmpsci}      % mathematics and physical sciences

%\bibliographystyle{ACM-Reference-Format}
\bibliography{bake}

\end{document}